\documentclass[9pt,a4paper,twocolumn,twoside]{rho-class/rho}
\usepackage[english]{babel}

\usepackage{booktabs}
\usepackage{threeparttable}
\usepackage{siunitx}
\usepackage{svg}
\usepackage{caption}
\usepackage{tikz}
\usetikzlibrary{arrows.meta, positioning, shapes.geometric, fit, backgrounds}

\newcommand{\etal}{\textit{et\penalty50\ al.}}

\title{Evidence-Grounded Frontier Mapping and Agentic Hypothesis Generation in Nanomedicine}
\author[{\textasteriskcentered,\textdagger},a]{Christiaan G.A. Viviers}
\author[{\textdagger},b]{Koen de Bruin}
\author[{\textdagger},b]{Mirre M. Trines}
\author[b]{Ayla M. Hokke}
\author[b]{ Roy van der Meel}
\author[c]{ Avi Schroeder}
\author[d]{ Twan Lammers}
\author[b,e]{Willem J.M. Mulder}
\author[a]{Fons van der Sommen}

\affil[a]{ARIA Lab, Signal Processing Systems, Department of Electrical Engineering, Eindhoven University of Technology, Eindhoven, The Netherlands.}
\affil[b]{Laboratory of Chemical Biology, Department of Biomedical Engineering, Eindhoven University of Technology, Eindhoven, The Netherlands.}
\affil[c]{The Louis Family Laboratory for Targeted Drug Delivery and Personalized Medicine Technologies, Department of Chemical Engineering, Technion - Israel Institute of Technology, Haifa, Israel.}
\affil[d]{Department of Nanomedicine and Theranostics, Institute for Experimental Molecular Imaging (ExMI), RWTH Aachen University Hospital, Aachen, Germany.}
\affil[e]{Department of Internal Medicine and Radboud Center for Infectious Diseases (RCI), Radboud University Medical Center, Nijmegen, The Netherlands.}

\corres{\textsuperscript{\textasteriskcentered}Corresponding author: \href{mailto:c.g.a.viviers@tue.nl}{c.g.a.viviers@tue.nl} (C. Viviers).}
\corres{\textsuperscript{\textdagger}These authors contributed equally to this work.}

\journalname{Arxiv}
\journal{Arxiv, May 2026}
\theday{\today}
\vol{X}
\no{X}

\begin{abstract}
    Nanomedicine research spans delivery chemistry, immunology, imaging, biomaterials, and disease-specific translational science, yet its conceptual design space remains fragmented across a large and heterogeneous literature. To date, artificial intelligence in nanomedicine has focused primarily on property prediction and formulation optimization, with much less attention to evidence-grounded discovery support at the level of research direction selection. We introduce \textit{pArticleMap}, a literature-mapping and research-hypothesis-generation system that combines article embeddings, similarity-graph analysis, sparse frontier extraction, structured evidence-pack retrieval, and an audited large-language-model~(LLM) workflow for grounded ideation. Rather than forecasting future concept co-occurrence, \textit{pArticleMap} targets low-density article-level bridge regions and cluster interfaces, then generates and scores citation-grounded hypotheses with large language models in an agentic setup. We evaluate the system with a retrospective realization benchmark (generate later literature under a historical cutoff) and a blinded human reader assessment layer across cue-conditioned nanomedicine tasks. Across 4 selected retrospective bundles, \textit{pArticleMap} generated ideas and selected task-retained hypotheses (winner ideas) under the benchmark protocol. Exact realization remained deliberately difficult, but under this benchmark \textit{pArticleMap} frequently reached the correct forward-looking literature neighborhood. For task-level retained hypotheses, a pooled gold recovery rate of 10.8\% was obtained, with a recall@10 of 15.9\% and a future-neighborhood rate of 61.0\%, indicating that the system often reached the correct forward-looking neighborhood~(paper ideas) even without exact paper-level recovery. Human--agent agreement is modest overall, indicating that internal scoring is useful as a support signal but does not replace expert judgment. These results position \textit{pArticleMap} as a conservative, evidence-grounded research assistant for nanomedicine\footnote{Code for pArticleMap is available \href{https://github.com/cviviers/nanotech_agent}{https://github.com/cviviers/nanotech\_agent}}.
\end{abstract}


\begin{document}

    \maketitle
    \thispagestyle{firststyle}


\section{Introduction}

\rhostart{N}anomedicine promises to connect advances in materials science, pharmacology, immunology, molecular imaging, and precision therapeutics. In practice, however, the field remains difficult to navigate at scale. The relevant literature is large, heterogeneous, and 

\begin{figure}[hb]
    \centering
    \includegraphics[width=1\linewidth]{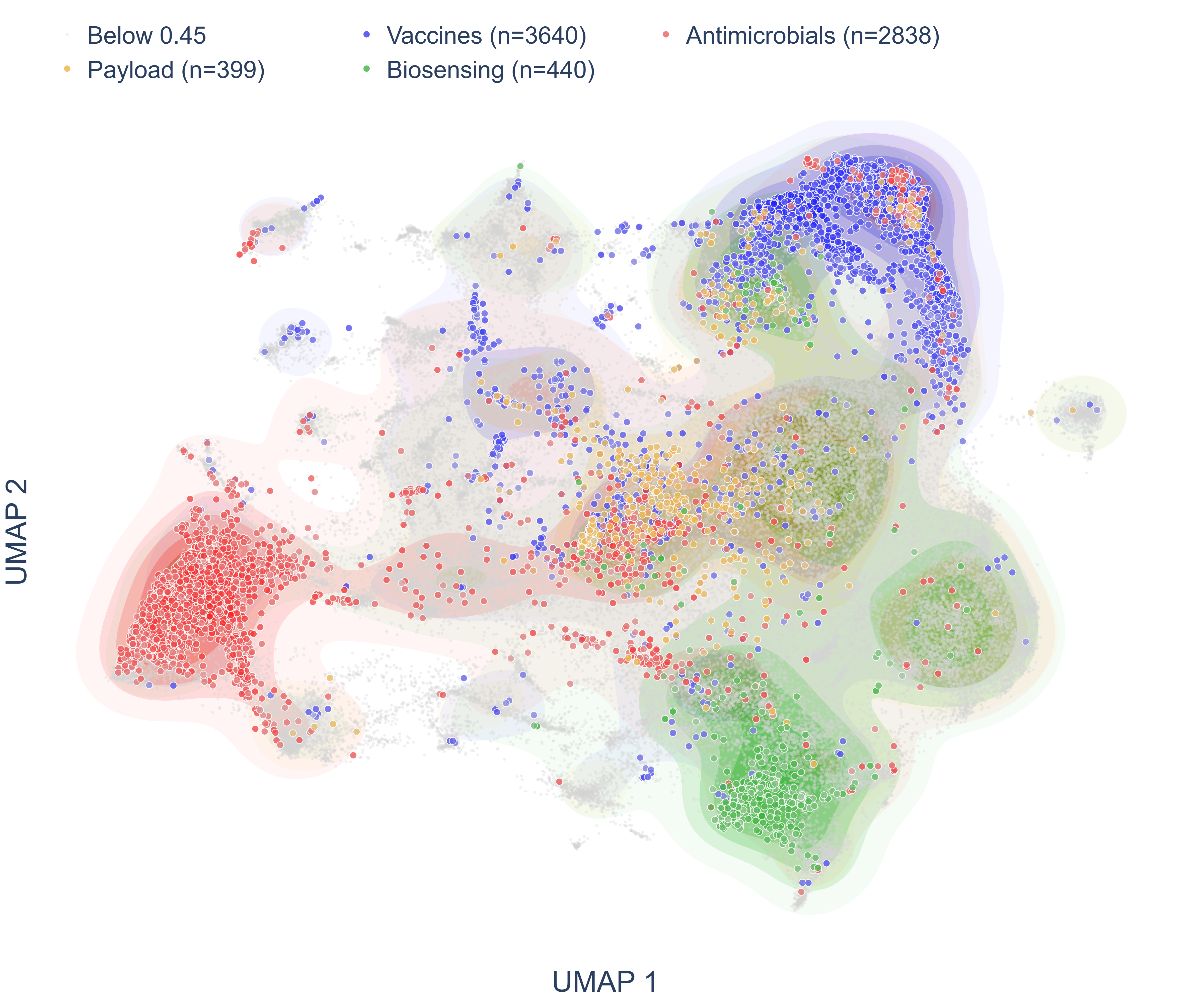}
    \caption{Similarity-weighted semantic overlays reveal how major nanotechnology themes occupy distinct but partially overlapping regions of the full-corpus embedding space.}
    \label{fig:semantic_overlay}
\end{figure}

\noindent fragmented across disease areas, delivery routes, material classes, and experimental models. As a result, potentially important conceptual connections often remain buried in separate subfields even when the underlying evidence already exists in public literature, despite long-standing calls to broaden nanomedicine beyond narrowly optimized delivery formulations~\cite{vandermeel2019smart, halwani2022development, anchordoquy2024mechanisms}.

This problem is especially acute in nanomedicine because the field is both scientifically broad and translationally constrained. Many research programs focus on incremental optimization within established formulation families, whereas clinically meaningful innovation often requires bridging distant domains: for example by combining established delivery concepts associated with one application with immunological or imaging strategies from another, or by importing enabling technologies from adjacent biomedical literatures. The challenge is therefore not only to search for similar papers, but to identify underexplored interfaces between partially separated research communities and to convert those interfaces into testable, evidence-backed hypotheses.

Recent progress in artificial intelligence offers several ingredients for addressing this problem. Scientific embedding models can represent large corpora in semantically meaningful vector spaces. Graph-based analysis and density estimation can expose corpus structure beyond keyword search. Retrieval-augmented language models can synthesize evidence across multiple papers. At the same time, experience from literature-based discovery and biomedical question answering shows that unconstrained generation is insufficient for scientific use: provenance, evidence traceability, and explicit handling of uncertainty are essential~\cite{swanson1986fish, cameron2015context, lewis2020retrieval}. A useful discovery system therefore requires not just a capable language model, but a full workflow that links corpus construction, representation learning, target identification, evidence retrieval, structured reasoning, and evaluation.

In parallel, AI in nanomedicine has so far been used primarily for tasks such as nanoparticle property prediction, formulation optimization, and high-throughput screening~\cite{serov2022artificial, li2024accelerating, ortizperez2024machine, hanna2026highthroughput}. These are valuable applications, but they address a narrower question than the one faced by researchers at the conceptual design stage. Before a formulation can be optimized, investigators must first decide which therapeutic direction is worth pursuing, which mechanistic bridge is plausible, and which combination of material, payload, targeting strategy, and disease context is genuinely underexplored. That earlier stage remains dependent on reading across a large and rapidly growing literature.

We address this gap with \textit{pArticleMap}, a human-centered system for evidence-grounded literature mapping and hypothesis generation in nanomedicine. \textit{pArticleMap} maps article-level literature structure to identify frontier regions and support grounded hypothesis generation. It represents PubMed-scale title-and-abstract corpora in a dense embedding space, organizes papers into literature communities, identifies low-density bridge regions between those communities, assembles structured evidence packs around selected targets, and uses an audited agentic workflow to generate bridge hypotheses and preclinical blueprints. \textit{pArticleMap} is specifically designed as a scientific assistant rather than an autonomous scientist: analysts retain control over corpus scope, analysis parameters, target selection, and post-generation review.

This article-level frontier-targeting perspective is important. Forecasting future concept co-occurrence in a concept graph is easier to benchmark, but it remains a coarse proxy for the real discovery-support task faced by nanomedicine researchers. \textit{pArticleMap} instead operates on article-level frontier regions, structured evidence packs, and grounded bridge hypotheses. That makes the task harder to evaluate, but more relevant to real research support because the outputs can be inspected against specific papers, evidence gaps, and later literature.

\begin{figure*}[ht]
    \centering
    \includegraphics[width=2.0\columnwidth]{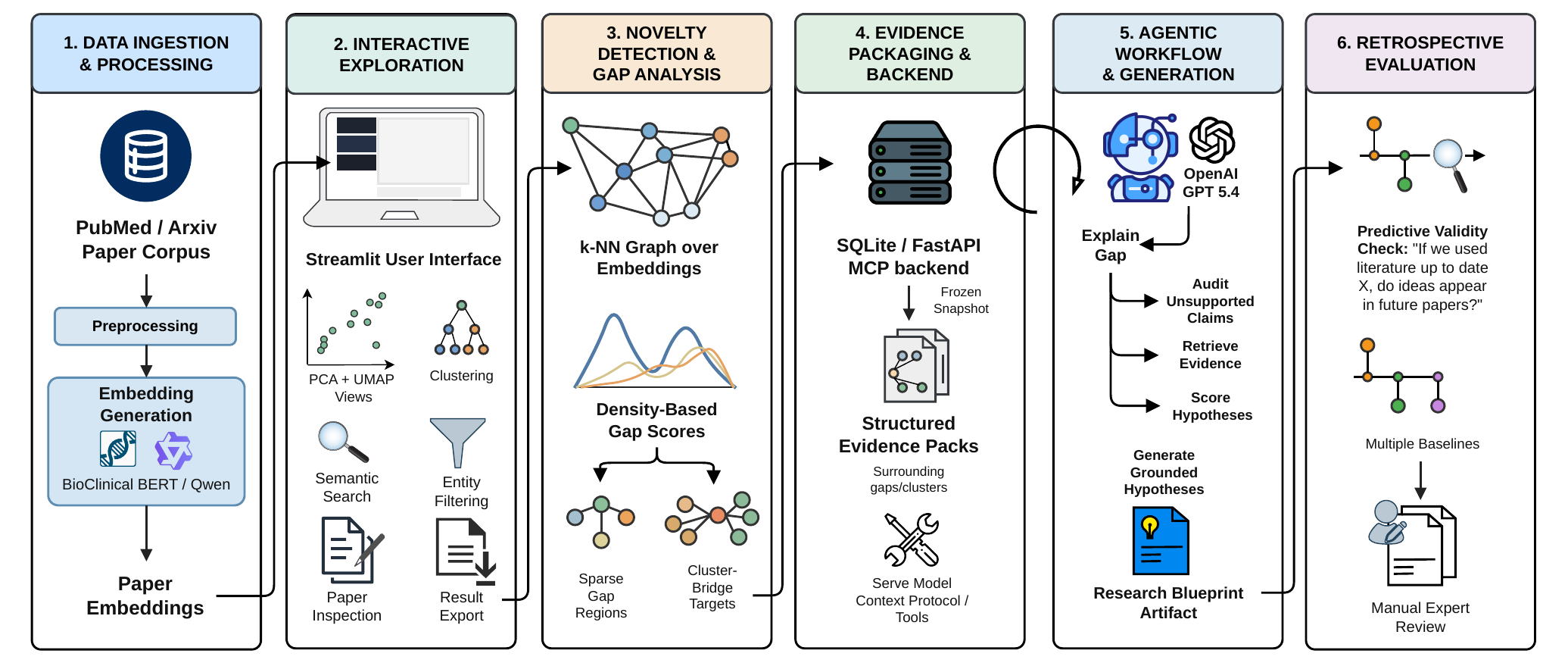}
    \caption{\textit{pArticleMap} maps article-level literature structure, surfaces sparse frontier regions, and uses evidence-constrained LLM workflows to generate and evaluate grounded research directions.}
    \label{fig:particemap-overview}
\end{figure*}

The contribution of this work is threefold. First, we present \textit{pArticleMap} as a reproducible representation-and-graph pipeline for identifying sparse, semantically coherent regions in the nanomedicine literature rather than relying on keyword gaps alone. Second, we introduce an evidence-pack-centered agentic workflow in which explanation, audit, retrieval patching, ideation, and blueprint generation are all tied to a literature snapshot and paper-level provenance. Third, we report a retrospective realization benchmark and a blinded human-review layer showing both the promise and the limits of internal groundedness scoring. Across four cue-conditioned retrospective bundles, \textit{pArticleMap} generated ideas and retained winner ideas; winner-level exact realization (generating an idea that was later published) remained difficult, but the system frequently reached the correct future neighborhood and exposed clear human--agent alignment, but also calibration gaps.

The remainder of the paper is organized as follows. We first situate \textit{pArticleMap} within literature-based discovery, retrieval-grounded scientific agents, and AI for nanomedicine. We then describe the article-level frontier-mapping and evidence-pack workflow, present the retrospective realization and human-review results, and close with limitations and implications for grounded scientific assistance.

\section{Related work}

\subsection{Research discovery}

The present work is grounded in the tradition of literature-based discovery~(LBD), which emerged from Swanson's demonstrations that non-obvious biomedical connections could be uncovered by linking disjoint literatures through shared intermediate concepts~\cite{swanson1986fish}. Classical LBD systems such as Arrowsmith~\cite{smalheiser1998arrowsmith} and BITOLA~\cite{hristovski2005bitola} operationalized this idea through term co-occurrence, controlled vocabularies, and interactive ranking interfaces. Their insight is that discovery support should expose intermediate evidence rather than return opaque scores.

Subsequent work moved from surface co-occurrence to semantic predications and knowledge graphs. Resources such as SemMedDB~\cite{kghubKGRegistry} enabled hypothesis generation over subject-predicate-object relations extracted from PubMed, while graph-based systems reframed discovery as link prediction or temporal forecasting. More recent methods such as MOLIERE~\cite{sybrandt2017moliere} and AGATHA~\cite{sybrandt2020agatha} combined large semantic graphs with learned rankers and historical holdout evaluation, bringing LBD closer to contemporary machine learning practice.

Large-scale biomedical knowledge graphs have also become practical substrates for AI-powered discovery. iKraph constructs a PubMed-scale biomedical knowledge graph for data-driven biomedical inference~\cite{zhang2025comprehensive}. At the same time, Scorpius shows that literature-derived medical knowledge graphs can be vulnerable to LLM-generated poisoning attacks, where a single malicious abstract can distort downstream drug--disease rankings~\cite{yang2024poisoning}. These results motivate our emphasis on temporal cutoffs, explicit provenance, and auditing of evidence packs rather than treating extracted literature structure as automatically trustworthy.

At the same time, several reviews have pointed out that LBD evaluation has often been methodologically weak. The field has relied heavily on a small number of canonical rediscoveries, making it difficult to assess generalization across domains or to separate meaningful foresight from trivial recovery of already implicit links. This criticism has motivated the use of larger temporal benchmarks, time-sliced discovery protocols, and more explicit leakage controls. Our retrospective benchmark is aligned with that direction: hypotheses are generated using only pre-cutoff evidence and evaluated only after generation against later literature.

Our work also relates to scientific corpus cartography. Embedding-based mapping, bibliometric visualization, and topic-model pipelines have shown that large research fields can be organized into interpretable structures using dimensionality reduction, clustering, and neighborhood graphs. Tools such as VOSviewer~\cite{vanEck2010vosviewer, vanEck2011textmining} and CiteSpace~\cite{chen2006citespace} established the value of interactive science maps for identifying communities, fronts, and pivots. \textit{pArticleMap} differs from traditional bibliometrics in using dense text embeddings as the primary similarity signal and in explicitly searching for sparse bridge regions between communities rather than only characterizing dense clusters or citation-based fronts.

This distinction also separates \textit{pArticleMap} from recent concept-graph forecasting systems. Science4Cast~\cite{Krenn2023} framed discovery as future-link prediction of concepts in an evolving semantic network of AI concepts, while Marwitz~\etal~\cite{marwitz2026predicting} used LLM-extracted materials-science concepts to build a concept graph and predict emerging concept combinations as candidate research directions. These works provide the closest benchmark-style comparators for our retrospective framing, but they operate primarily at the concept-pair level. \textit{pArticleMap} instead operates on sparse frontier regions in article space, then builds structured evidence packs and grounded bridge hypotheses around those regions. The resulting task is harder to score cleanly, but closer to how researchers actually inspect literature, compare evidence, and formulate new directions.

\subsection{Agents}
Recent scientific question-answering and literature-synthesis systems have shown that retrieval-augmented generation is the default starting point for grounded language-model reasoning in research settings. The key idea, formalized in modern retrieval-augmented generation~(RAG), is to condition a parametric generator on non-parametric evidence retrieved at inference time rather than relying on model weights alone~\cite{lewis2020retrieval}. In biomedicine, this design is particularly important because the literature is dynamic, terminology is specialized, and unsupported claims can be difficult to detect without explicit provenance.

Several recent systems provide relevant precedents. PaperQA~\cite{lala2023paperqa} and PaperQA2~\cite{skarlinski2024paperqa2} frame scientific question answering as an agentic literature workflow with retrieval, reranking, evidence summarization, and explicit citation handling. OpenScholar~\cite{asai2026openscholar} extends this line toward long-form literature synthesis and treats citation accuracy as a primary evaluation target rather than a secondary attribute. Across these systems, the dominant design pattern is multi-stage retrieval plus evidence transformation, not simple top-$k$ concatenation.

Recent scientific LLM systems illustrate adjacent design patterns. ChemCrow couples an LLM agent to chemistry tools for synthesis planning, molecular design, and expert-evaluated chemical reasoning~\cite{bran2024augmenting}. LLM4SD uses LLMs to synthesize literature knowledge and infer data-derived molecular-property rules that become interpretable model features~\cite{zheng2025large}. PhyE2E combines neural models, symbolic search, and LLM-generated synthetic expressions for formula discovery in space physics~\cite{ying2025neural}. These systems pursue downstream domain tasks, whereas \textit{pArticleMap} uses an agent to interrogate a corpus boundary and produce evidence-grounded hypotheses for human consideration.

\textit{pArticleMap} adopts the same general philosophy but differs in objective and target construction. PaperQA-style systems are optimized primarily for answering user questions from retrieved literature. In contrast, \textit{pArticleMap} begins with structural targets derived from the corpus itself: low-density gap regions and cluster-pair interfaces in an embedding-space graph. The role of the agent is therefore not to answer an arbitrary question, but to explain a sparse literature boundary, audit the sufficiency of evidence, and generate bridge hypotheses that are explicitly tied to that boundary.

The workflow also relates to emerging multi-agent and self-correcting scientific assistants. Recent surveys of LLM-based hypothesis generation~\cite{alkan2025hypothesis_survey, kulkarni2025hypothesis_validation, ren2025scientific_agents, zheng2025automation_autonomy} emphasize retrieval grounding, iterative refinement, and critique or verification loops as key mechanisms for reducing hallucination. The explain-audit-patch-retrieve-ideate pattern in \textit{pArticleMap} is a concrete instantiation of that design principle. The audit stage is especially important scientifically because it transforms unsupported claims into explicit retrieval actions instead of allowing ideation to proceed over weakly grounded context.

Finally, recent large-scale work on LLM feedback in peer review shows the level of statistical and human-evaluation rigor now expected when LLM systems are inserted into scientific workflows~\cite{thakkar2026large}. We therefore frame the human layer in \textit{pArticleMap} as a calibration and validation problem, not as anecdotal endorsement of generated ideas.

\subsection{Nanomedicine}

Within nanomedicine itself, most AI applications have focused on formulation design, physicochemical property prediction, biodistribution modeling, or high-throughput optimization. This includes machine-learning-guided lipid discovery, nanoparticle design workflows, and predictive models for delivery performance~\cite{serov2022artificial, li2024accelerating, ortizperez2024machine, hanna2026highthroughput}. These efforts are important, but they typically assume that a target therapeutic direction has already been chosen. They do not directly address the earlier-stage problem of identifying underexplored conceptual opportunities across a broad literature landscape.

That omission matters because nanomedicine is unusually sensitive to cross-domain transfer. Useful ideas may arise when delivery strategies, material platforms, immune mechanisms, diagnostic concepts, or disease models migrate across subfields. Yet the relevant literature is fragmented by terminology, assay conventions, and translational context. The same scientific concept may be expressed through different material names, targeting ligands, or disease-specific language, making purely lexical search insufficient.

Nanomedicine also brings domain-specific constraints that are underemphasized in generic AI-for-science systems. Terminological harmonization remains a persistent issue, as shown by work on regulatory nanomedicine terminology and by ontology efforts such as the NanoParticle Ontology. More broadly, clinically meaningful nanomedicine innovation depends not only on novelty but also on feasibility, safety, manufacturability, and biological context. Nanosafety infrastructures and FAIR data efforts illustrate that relevant evidence often extends beyond unstructured papers alone.

\textit{pArticleMap} is positioned as complementary to, not competitive with, formulation-level machine learning. We aim to support the conceptual stage of nanomedicine research by helping researchers navigate a large literature, identify sparse interfaces between partially disconnected topics, and generate hypotheses that remain tied to explicit evidence. This framing is intentionally conservative: \textit{pArticleMap} is a discovery-support tool for researchers, not an autonomous platform for optimizing nanomedicine end to end.

\section{Methods}\label{methods}

We formulate \textit{pArticleMap} as a human-directed literature-mapping and evidence-grounded hypothesis-generation system rather than as an autonomous scientific discovery engine. Let

\begin{equation}
    \mathcal{D} = \{(d_i, m_i)\}_{i=1}^{N}
\end{equation}

denote a corpus of scientific papers, where $d_i$ is the title-plus-abstract text of paper $i$ and $m_i$ is its associated metadata. The implemented workflow maps this corpus to

\begin{equation}
    \Phi : \mathcal{D} \rightarrow (\mathcal{R}, \mathcal{S}, \mathcal{E}, \mathcal{H}, \mathcal{B}),
\end{equation}

where $\mathcal{R}$ is a set of sparse literature regions, $\mathcal{S}$ is a frozen analysis snapshot, $\mathcal{E}$ is a target-specific evidence pack, $\mathcal{H}$ is a set of grounded bridge hypotheses, and $\mathcal{B}$ is an optional preclinical blueprint for the top-scored hypothesis. Analysts retain control over corpus scope, analysis configuration, target selection, and downstream review throughout.

This section describes the core \textit{pArticleMap} workflow used for frontier mapping, evidence construction, and grounded ideation. The retrospective benchmark that reuses this workflow is described separately in Section~\ref{setup}.

\subsection{Data collection \& representation}

The current \textit{pArticleMap} implementation operates on a PubMed-derived JSON corpus built from yearly queries around nanoparticle and in vivo nanomedicine literature. Each record preserves bibliographic metadata together with title, abstract, DOI when available, author keywords, MeSH terms, language information, and publication-date components. The full data collection and conversion scripts are available on Github~\footnote{\textit{pArticleMap} code \href{https://github.com/cviviers/nanotech_agent}{https://github.com/cviviers/nanotech\_agent}.}

For downstream analysis, each paper is represented by the concatenation of title and abstract. This is a deliberate design choice: abstracts are broadly available and standardized across the corpus, whereas full text is sparse, heterogeneous, and difficult to use reproducibly at corpus scale. Documents are embedded by an encoder

\begin{equation}
f_{\mathrm{emb}} : d_i \mapsto x_i \in \mathbb{R}^{p}.
\end{equation}

The system currently supports two normalized embeddings obtained from the large embedding models `Qwen3-Embedding-0.6B`~\cite{zhang2025qwen3embeddingadvancingtext} and `BioClinical-ModernBERT-large`~\cite{sounack2025bioclinicalmodernbertstateoftheartlongcontext}. The selected embedding set becomes the primary representation for both corpus analysis and, when stored in the snapshot, downstream retrieval. In Appendix~\ref{appendix:text-encoder-eval} we provide in-depth analysis of these embedding models and conclude that Qwen-3 is the optimal text encoder for this system.

\textit{pArticleMap} distinguishes between the primary embedding space and the analysis space. Let $x_i$ denote the primary embedding. When PCA is enabled, the analysis operates on

\begin{equation}
z_i = P_r(x_i) \in \mathbb{R}^{r},
\end{equation}

where $P_r$ is a PCA projection retaining the top $r$ components. In the canonical headless configuration, PCA is enabled by default. This is implemented to improve overall system speed as running subsequent analysis on the large 1024 embedding space can be computationally expensive. UMAP projections are also computed, but they are used only for visualization and are not part of gap scoring or target ranking (discussed in Sections~\ref{sec:cluster} and \ref{sec:gap}).

\subsection{Similarity graph, density measure \& clustering}~\label{sec:cluster}

\subsubsection{Operational clustering} 

The reusable analysis pipeline in \textit{pArticleMap} produces one operational clustering that is carried forward into snapshot construction and evidence retrieval. Three clustering modes are supported: hdbscan~\cite{Malzer_2020_hdbscan}, K-means~\cite{lloyd1982least} and a graph-community Leiden algorithm~\cite{Traag_2019_leiden}.

The pipeline also supports a graph-community mode in which a $k$-nearest-neighbor graph is built first and community labels are obtained with Leiden or Louvain. Regardless of which option is used, only one selected cluster assignment is propagated downstream.

To characterize local literature density, the pipeline constructs an undirected $k$-nearest-neighbor graph on the analysis vectors using cosine distance

\begin{equation}
\delta(u,v) = 1 - \frac{u^\top v}{\|u\|_2\|v\|_2}.
\end{equation}

With default graph size $k=21$, an edge $(i,j)$ is added when $j$ appears in the nearest-neighbor set of $i$. The stored edge weight is cosine similarity,

\begin{equation}
w_{ij} = 1 - \delta(z_i, z_j).
\end{equation}

and when an undirected edge is observed from both directions, the larger weight is retained. Neighbor search is executed in batches for memory stability but does not change the graph definition.

Paper-level sparsity is then estimated across several neighborhood scales. For each $k \in \mathcal{K}$ (default $\mathcal{K}  = \{10,20,30,40,50\}$, the pipeline computes the mean $k$-neighbor distance

\begin{equation}
\rho_i^{(k)} = \frac{1}{k}\sum_{j \in \mathcal{N}_k(i)} \delta(z_i, z_j),
\end{equation}

standardizes that quantity corpus-wide,

\begin{equation}
\tilde{\rho}_i^{(k)} = \frac{\rho_i^{(k)} - \mu_k}{\sigma_k},
\end{equation}

and defines the final paper-level gap score as

\begin{equation}
g_i = \frac{1}{|\mathcal{K}|}\sum_{k \in \mathcal{K}} \tilde{\rho}_i^{(k)}.
\end{equation}

This is the implemented novelty proxy: papers with high $g_i$ remain relatively isolated across several neighborhood scales rather than under a single arbitrary choice of $k$.

\subsection{Gap region extraction \& bridge targets}\label{sec:gap}

Candidate gap papers are defined by a high quantile of the empirical gap-score distribution. Let $q_\tau$ be the $\tau$-quantile of $\{g_i\}_{i=1}^{N}$, with default $\tau = 0.95$. The candidate set is

\begin{equation}
\mathcal{C}_\tau = \{i : g_i \ge q_\tau\}.
\end{equation}

The induced subgraph

\begin{equation}
G_\tau = G[\mathcal{C}_\tau]
\end{equation}

is decomposed into connected components, and only components of size at least $m$ are kept, with default $m=3$. Each retained component becomes a gap region. In the analysis stage these regions are sorted by size; during snapshot-backed retrieval they are surfaced through `top gaps`, which ranks stored gaps by average gap score and then size.

For each gap region $R_r$, the snapshot stores region membership, paper-level ranks, summary statistics, and the set of operational clusters it touches,

\begin{equation}
\mathcal{C}(R_r) = \{c_i : i \in R_r,\; c_i \neq -1\}.
\end{equation}

These touched-cluster sets support a second target type besides the gap itself. Cluster-pair bridge targets are constructed from unordered pairs of clusters that co-occur around top-ranked gaps. If this does not yield enough pairs, \textit{pArticleMap} backs off to pairwise combinations among the largest clusters in the snapshot. As a result, downstream generation can target either a sparse region directly or the interface between two denser literature communities.

\subsection{Evidence-pack construction \& publication}\label{sec:evidence_pack}

After analysis, the corpus state is frozen into a reusable snapshot and published to the backend. A snapshot contains paper records, embeddings, cluster assignments, gap-region summaries, gap-to-paper membership, and analysis metadata such as embedding choice and clustering configuration. The backend is implemented as a FastAPI service over a SQLite knowledge store. These API endpoints are set up as the Model Context Protocol~(MCP)~\cite{mcp_spec, anthropic_mcp_2024}. MCP is an open protocol for connecting LLM applications to external tools and data sources. In this case, it is used to standardize how language models construct the evidence pack used for hypothesis generation. This improves interoperability, reuse, and maintainability of \textit{pArticleMap}.

\begin{rhoenv}[frametitle=Example Evidence Pack]
\textbf{Cue.} ``What adjuvants can be included in mRNA LNP vaccines to improve their long term efficacy?'' \\
\textbf{Target.} \texttt{cluster pair 0-11} induced by \texttt{gap 4}; the published evidence pack contained 64 papers. \\
\textbf{Key support.} The retained hypothesis later cited five papers from this pack: \emph{Supramolecular Hydrogel from Nanoparticles and Cyclodextrins for Local and Sustained Nanoparticle Delivery} (2016), \emph{Site-Specific Construction of Long-Term Drug Depot for Suppression of Tumor Recurrence} (2019), \emph{A magnetic chitosan hydrogel for sustained and prolonged delivery of Bacillus Calmette--Gu\'erin in the treatment of bladder cancer} (2013), \emph{Injectable methylcellulose hydrogel containing calcium phosphate nanoparticles for bone regeneration} (2018), and \emph{Incorporation of BMP-2 nanoparticles on the surface of a 3D-printed hydroxyapatite scaffold using an $\varepsilon$-polycaprolactone polymer emulsion coating method for bone tissue engineering} (2018). \\
\textbf{Interpretation.} This pack bridges depot-release and scaffold/hydrogel literatures, but it does not directly establish which adjuvants should be incorporated into mRNA-LNP vaccines to improve long-term efficacy. A full worked example appears in Appendix~\ref{appendix:worked-vaccine}.
\end{rhoenv}

This snapshot boundary is central to \textit{pArticleMap}. Once a snapshot has been published, all downstream retrieval and generation run against that immutable snapshot identifier rather than against mutable in-memory analysis state. The same backend is reused by the Streamlit app (User Interface), the interactive CLI, and the retrospective evaluation runner.

Given a target $\tau$ and a retrieval budget, a structured evidence pack is constructed as

\begin{equation}
\mathcal{E}_\tau =
\mathcal{E}_{\mathrm{exemplar}}
\cup
\mathcal{E}_{\mathrm{boundary}}
\cup
\mathcal{E}_{\mathrm{gap}}
\cup
\mathcal{E}_{\mathrm{diverse}}
\cup
\mathcal{E}_{\mathrm{query}},
\end{equation}

where every paper is annotated with selection provenance such as selection sources and selection meta data.

For cluster exemplars, if embeddings are available in the snapshot, \textit{pArticleMap} computes a cluster centroid

\begin{equation}
\mu_c = \frac{1}{|S_c|}\sum_{i \in S_c} x_i
\end{equation}

and chooses papers nearest to that centroid under cosine distance. If embeddings are unavailable, it falls back to a deterministic SQL ordering.

For a `gap` target, the evidence pack starts with the highest-ranked papers already assigned to that gap region and then supplements them with exemplar papers from the clusters touched by the gap.

For a `cluster pair` target $(a,b)$, the backend retrieves exemplars from both clusters and then searches for papers near the boundary between their centroids. For a paper $i \in S_a$, \textit{pArticleMap} computes

\begin{equation}
d_a(i)=\delta(x_i,\mu_a), \qquad d_b(i)=\delta(x_i,\mu_b),
\end{equation}

a boundary margin

\begin{equation}
m_i = |d_a(i)-d_b(i)|,
\end{equation}

and a midpoint-closeness term

\begin{equation}
s_i = \frac{1}{2}\big(d_a(i)+d_b(i)\big).
\end{equation}

Papers are ordered lexicographically by $(m_i, s_i, d_a(i))$, which prioritizes records close to both cluster prototypes and near the decision boundary between them. If boundary slots remain, the pack can be supplemented with papers from gap regions that touch both clusters. When requested, additional diverse papers and lexical query matches can also be added.

The backend also accepts an optional structured discovery cue. The cue is used only for steering: it can generate extra retrieval queries, attach cue-alignment metadata to selected papers, and rerank the final pack toward cue-relevant evidence. It is never treated as literature evidence and must not be cited as support.

\subsection{Agent orchestration}\label{sec:agent}

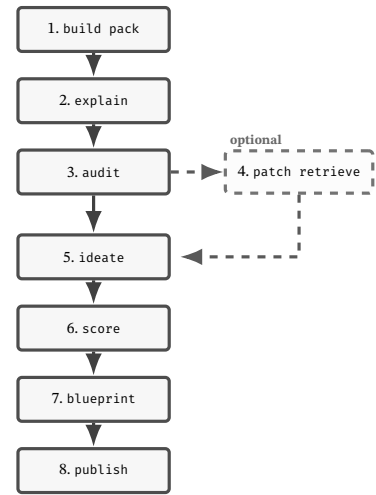
\begin{figure}[ht]
    \centering
    \begin{tikzpicture}[
        font=\scriptsize,
        >=Latex,
        node distance=3.2mm and 6mm,
        stage/.style={
            rectangle,
            rounded corners=1.5pt,
            draw=black!70,
            very thick,
            minimum width=2.0cm,
            minimum height=5.8mm,
            inner sep=1.2pt,
            align=center,
            fill=black!3
        },
        optstage/.style={
            rectangle,
            rounded corners=1.5pt,
            draw=black!55,
            dashed,
            very thick,
            minimum width=1.95cm,
            minimum height=5.6mm,
            inner sep=1.2pt,
            align=center,
            fill=black!1
        },
        line/.style={->, very thick, draw=black!75},
        optline/.style={->, very thick, dashed, draw=black!65},
        labelbox/.style={
            font=\tiny\bfseries,
            text=black!60,
            inner sep=0.8pt
        }
    ]

    \node[stage] (build) {1.\ \texttt{build pack}};
    \node[stage, below=of build] (explain) {2.\ \texttt{explain}};
    \node[stage, below=of explain] (audit) {3.\ \texttt{audit}};
    \node[stage, below=5mm of audit] (ideate) {5.\ \texttt{ideate}};
    \node[stage, below=of ideate] (score) {6.\ \texttt{score}};
    \node[stage, below=of score] (blueprint) {7.\ \texttt{blueprint}};
    \node[stage, below=of blueprint] (publish) {8.\ \texttt{publish}};

    \node[optstage, right=7mm of audit] (patch) {4.\ \texttt{patch retrieve}};

    \draw[line] (build) -- (explain);
    \draw[line] (explain) -- (audit);
    \draw[line] (audit) -- (ideate);
    \draw[line] (ideate) -- (score);
    \draw[line] (score) -- (blueprint);
    \draw[line] (blueprint) -- (publish);

    \draw[optline] (audit.east) -- (patch.west);
    \draw[optline] (patch.south) |- ([xshift=1.2mm]ideate.east);

    \node[labelbox, above right=0mm and 0.5mm of patch.north west] {optional};

    \end{tikzpicture}
    \caption{Implemented agent workflow. After \texttt{audit}, the pipeline may optionally invoke \texttt{patch retrieve} before continuing to \texttt{ideate}, \texttt{score}, \texttt{blueprint}, and \texttt{publish}. A detailed schematic of the agent orchestration is available in Figure~\ref{appendix:agent-workflow}. }
    \label{fig:agent-workflow}
\end{figure}

The canonical \textit{pArticleMap} generation path is implemented as a LangGraph~\cite{langchainLangGraphAgent} state machine over a target-specific snapshot-backed state

\begin{equation}
s_t = (\tau, \mathcal{E}_t, X_t, A_t, H_t, Q_t, B_t, u_t),
\end{equation}

where $\tau$ is the target, $\mathcal{E}_t$ is the evidence pack, $X_t$ is a contrastive explanation, $A_t$ is an audit report, $H_t$ is a set of generated hypotheses, $Q_t$ is a set of idea-quality scores, $B_t$ is a blueprint, and $u_t$ is the number of retrieval-patching iterations completed so far. The implemented node order is depicted in Figure~\ref{fig:agent-workflow}.

During `explain`, the language model receives the serialized evidence pack and produces a structured contrastive account of the target. For gap targets it is asked to explain what lies on either side of the sparse region; for cluster-pair targets it is asked to explain why the two communities remain separated in embedding space. The required output includes side summaries, axes of separation, bridge seeds, and explicit insufficient-evidence signaling.

During \textit{audit}, the model reviews the explanation for unsupported claims, missing facets, cue violations, and candidate patch-retrieval queries. If the audit reports that more evidence is needed, or if cue alignment is judged too weak when a cue is active, the workflow enters `patch\_retrieve`. That node performs another focused retrieval pass using the audit-generated lexical queries and merges newly retrieved papers into the evidence pack. The explanation and audit loop in \textit{pArticleMap} is bounded by a configurable iteration budget.

Once the pack is considered sufficient, `ideate` generates structured bridge hypotheses grounded in the current evidence pack and contrastive explanation. `score` then evaluates the generated ideas on six dimensions: importance, novelty, plausibility, feasibility, evaluability, and likely impact. When an OpenAI-backed evaluator is available this step uses a structured LLM judge; otherwise it falls back to a deterministic heuristic scorer so the pipeline still runs end to end.

`blueprint` selects the top-scored hypothesis and expands it into a concise preclinical plan covering materials, synthesis and characterization, in vitro work, in vivo work, risks, mitigations, and success criteria. Finally, `publish` stores the full research brief in the backend, including evidence metadata, explanation, audit output, hypotheses, idea scores, blueprint, and iteration count.

Across all nodes, prompts require the model to stay within the retrieved evidence pack, cite by paper ID, and mark unsupported details as assumptions or unknowns. \textit{pArticleMap} is therefore best understood as an auditable scientific-assistance workflow: it helps users navigate sparse interfaces in the nanomedicine literature, but it does not validate the resulting ideas experimentally and does not replace scientific judgment.

\section{Experimental Setup \& Validation Protocol}\label{setup}

\subsection{Retrospective validation}\label{retro_validation}
We evaluate \textit{pArticleMap} with a temporally held-out retrospective benchmark.. The objective is not to claim autonomous scientific discovery, but to test a narrower and more defensible question: if the system is restricted to literature available before a historical cutoff, can the system produce historically grounded hypotheses that later retrieve specific future papers or at least land in the correct forward-looking literature neighborhood?

The benchmark reuses the same analysis, snapshot, evidence-pack, and generation stack described in Section~\ref{methods}, but enforces a temporal split over the full cleaned PubMed-derived corpus. In the default configuration, the historical corpus contains papers published on or before \texttt{2019-12-31}, while the primary future evaluation window spans \texttt{2020-01-01} to \texttt{2026-01-01}. When publication month or day is missing, dates are conservatively imputed to the last day of the most specific known period so that uncertain records are not accidentally moved into the historical split.

Historical analysis is then performed only on the pre-cutoff corpus using the shared headless configuration: Qwen embeddings as the default analysis store, PCA with $102$ components, Leiden as the primary clustering method with implemented fallback to K-means when needed, a $21$-nearest-neighbor similarity graph, density scales~$\{10,20,30,40,50\}$, gap quantile~$0.95$ and minimum gap-region size~$3$. The resulting analyzed corpus is serialized into an immutable historical snapshot and published to the backend. All downstream evidence retrieval, target selection, and generation operate only on this frozen historical snapshot, which is the main leakage-control mechanism in the evaluation.

From the historical snapshot, the runner selects two classes of frontier targets: top-ranked gap regions and cluster-pair bridge targets. In the canonical benchmark, up to $20$ gap targets and $10$ cluster-pair targets are retained. For each target, a focused evidence pack is built with a narrow retrieval budget (\texttt{exemplars = 8}, \texttt{boundary = 8}, \texttt{diverse = 0}) so that the benchmark remains centered on the local frontier rather than on broad background retrieval. Future papers are not scored indiscriminately. Instead, each future paper is first screened for strong historical near-duplicates, converted into a pseudo-cue, and assigned to the historically most compatible frontier target. The benchmark then retains the top~$50$ target-assignable future papers as \text{gold recovery tasks}.

The evaluation runner can also compare the full agentic orchestrator against simpler baselines that operate on the same historical targets, including \texttt{single shot llm}, \texttt{retrieval summary direct}, \texttt{heuristic bridge}, \texttt{pack query baseline}, and \texttt{random target control}. Details about these are provided in Appendix~\ref{appendix:expected_outcome}. The full orchestrator follows the implemented sequence shown in Figure~\ref{fig:agent-workflow} and Figure~\ref{appendix:agent-workflow}, whereas the baselines remove parts of the audit and retrieval-refinement loop or replace target assignment with simpler controls. By default, each method is evaluated with up to $3$ hypotheses per target. The present work, however, focuses on the completed \textit{pArticleMap}-centered realization and calibration analyses rather than a full cross-method comparison table.


Unless otherwise stated, the main-text realization metrics use a task-level best-of-N retention step: for each evaluation task, we retain the generated hypothesis with the strongest gold-paper reciprocal-rank signal, using cue-weighted reciprocal rank when a discovery cue is active and raw reciprocal rank otherwise. This is an evaluation-time upper-bound estimate of task recoverability under the current generator, not a deployable internal ranking policy. Where the paper discusses practical ranking behavior, we refer instead to the model's internal idea scores.

\paragraph{Cue-conditioned benchmark tasks.}
To test \textit{pArticleMap} under domain-directed discovery cues rather than only generic frontier recovery, we constructed four additional semantically filtered text corpora from the master dataset. Each auxiliary corpus was created by applying a semantic filter using the instruction \emph{``Given these keywords, retrieve relevant passages related to the keywords''} together with a domain-specific keyword set. The corresponding natural-language research question was then provided to the generator as the discovery cue during retrospective evaluation. The four cue-conditioned tasks were:

\begin{itemize}
    \itemsep1em
    \item \textbf{Vaccines}\\
    \emph{keywords}: vaccine, adjuvants, mRNA lipid nanoparticles, ionizable lipid, antibody titer, immune memory, immunity;\\
    \emph{prompt question}: What adjuvants can be included in mRNA LNP vaccines to improve their long term efficacy?
    \item \textbf{Antimicrobials} \\
    \emph{keywords}: bacteria, antimicrobial, inorganic nanoparticles, coating, biofilm, surface, infection; \\
    \emph{prompt question}: What characteristics should a coating for inorganic nanoparticles have to overcome biofilms?
    \item \textbf{Payload integration} \\
    \emph{keywords}: payload, incorporation, functionalization, protein, delivery, multifunction;\\
    \emph{prompt question}: What are innovative strategies to incorporate both protein payloads and nucleic acids in the same nanoparticle platform?
    \item \textbf{Biosensing}\\
    \emph{keywords}: co-detection, sensing, biosensor, dual-target, DNA, RNA, nucleic acids, nanosensor; \\
    \emph{prompt question}: I'm looking for a nanoparticle biosensing system that can simultaneously detect DNA and RNA.
\end{itemize}

These cue-conditioned corpora serve two purposes. First, they restrict the future-paper pool to a scientifically coherent thematic slice, reducing the chance that a recovered paper is merely topically adjacent to the intended task. Second, they test whether cue-aware evidence retrieval and generation can remain historically grounded while still steering toward specific nanomedicine design questions.

\paragraph{Evaluation criteria.}
Each generated hypothesis is converted into a structured fingerprint spanning fields such as disease, material, payload, targeting strategy, mechanism, model, route, and intended outcome. That fingerprint is used to retrieve candidates separately from the historical corpus and the future corpus through a hybrid lexical, embedding-based, and reranking pipeline. Candidate matches are then assigned a deterministic match label using a combined score

\begin{equation}
    s = 0.55\, s_{\mathrm{rank}} + 0.25\, s_{\mathrm{field}} + 0.20\, \frac{s_{\mathrm{emb}} + 1}{2},
\end{equation}

where $s_{\mathrm{rank}}$ is the reranker score, $s_{\mathrm{field}}$ is weighted fingerprint-field overlap, and $s_{\mathrm{emb}}$ is embedding similarity. A candidate is then labeled according to Table~\ref{tab:match-criteria}. The respective weights and ranges were empirically chosen.

\begin{table}[t]
\centering
\small
\caption{Match categorization based on score thresholds.}
\begin{tabular}{l l}
\toprule
Category & Condition \\
\midrule
\texttt{strong match} & $s_{\mathrm{rank}}\geq0.80$, $s_{\mathrm{field}}\geq0.45$, $s\geq0.70$ \\
\texttt{partial match} & $s_{\mathrm{rank}}\geq0.58$, $s_{\mathrm{field}}\geq0.22$, $s\geq0.50$ \\
\texttt{background only} & $s_{\mathrm{rank}}\geq0.38$ or $s_{\mathrm{field}}\geq0.15$ \\
\texttt{no match} & otherwise \\
\bottomrule
\end{tabular}
\label{tab:match-criteria}
\end{table}

For each hypothesis, the benchmark records three quantities: the best historical candidate and its match label, the one-indexed rank $r$ of the assigned gold future paper in the future-corpus retrieval list, and the best future candidate other than the gold paper. If the gold paper is not retrieved, $r$ is treated as missing and its reciprocal rank is set to zero. Recovery labels are assigned with the following precedence. A hypothesis is labeled \texttt{historical confound} if the best historical candidate is a \texttt{strong match}, even if the gold paper is also retrieved. Otherwise it is labeled \texttt{gold recovered} if $r\leq10$, \texttt{future neighbor only} if the gold paper is not in the top ten but the best non-gold future candidate has any non-\texttt{no match} label, and \texttt{not recovered} otherwise. Thus \texttt{gold recovered rate} is stricter than \texttt{gold\_recall@10}: the former excludes cases that are better explained as rediscovery of pre-cutoff literature.

Aggregate metrics are computed after collapsing multiple generated hypotheses for the same \texttt{(method, seed, gold future paper)} task to a single task-level row. The selected row is the hypothesis with the largest reciprocal-rank signal, using cue-weighted reciprocal rank when a discovery cue is active and raw reciprocal rank otherwise, with hit indicators and mean idea score used only as tie-breakers. For $N$ task-level rows, 
\begin{equation}
\texttt{gold recall@k} = N^{-1}\sum_i 1[r_i\leq k]\, \text{for}\,  k\in\{1,5,10\},
\end{equation}
and 
\begin{equation}
\texttt{gold MRR} = N^{-1}\sum_i 1/r_i,\, \text{with}\, 1/r_i=0
\end{equation}
when the gold paper is absent. The label rates are simple proportions over the same $N$ rows: \texttt{gold recovered rate}, \texttt{historical confound rate}, and \texttt{future neighbor only rate} count rows with the corresponding recovery label and divide by $N$. When a cue is active, we additionally report cue-weighted recall and reciprocal-rank variants so that topical relevance and future-paper recovery can be evaluated jointly.

A \texttt{future neighbor only} outcome indicates that, after excluding strong historical confounds, the retained hypothesis did not recover the assigned gold paper in the top ten but did retrieve a semantically matched non-gold future paper, suggesting neighborhood-level anticipation rather than exact paper-level realization.

This retrospective protocol provides a stronger validation signal than anecdotal rediscovery because it enforces historical grounding, explicit leakage checks, fixed target selection, and held-out future evaluation. At the same time, it remains a proxy benchmark: recovering a later paper does not by itself establish mechanistic correctness, translational value, or clinical importance. For that reason, the automated benchmark for \textit{pArticleMap} is paired with manual expert review.

\subsection{Human evaluation}\label{human_validation}
The human-evaluation layer is designed as the paper-grade adjudication step for ambiguous automated matches. For each retrospective run, we export review packets built from the best hypothesis for each \texttt{(method, seed, gold future paper)} task. Each review row contains the assigned historical target, the generated hypothesis, idea-quality scores (kept hidden during evaluation), a compact evidence-pack summary, the top historical retrievals, the top future retrievals (also hidden during evaluation), and any cue metadata used during generation.

Reviewers are tasked to perform a blind-first evaluation of agent-generated nanomedicine hypotheses. Each reviewer is presented with the generated hypothesis text, discovery cue, reasoning and audit metadata, and the associated evidence pack, then prompted to score each idea on a five-point scale for importance, novelty, plausibility, feasibility, evaluability, and likely impact. Reviewers can add criterion-specific notes, an overall rationale, confidence level, and flags for insufficient context or need for expert follow-up; per-criterion model scores and retrospective retrieval outcomes remained hidden until the reviewer submitted the assessment.

Expert reviewers can therefore inspect not only whether a future paper was retrieved, but also whether the generated idea is genuinely novel at the cutoff date, semantically aligned with the matched future paper, plausible from a nanomedicine perspective, and useful as a research direction. In practice, this review layer is necessary because the automatic labels rely on heuristic fingerprint extraction and reranking thresholds. A strong automated future match may still reflect only broad topical overlap, while a meaningful bridge hypothesis may be undercounted if it is phrased differently from later literature.

Accordingly, we treat automated retrospective recovery as a scalable screening layer and expert review as the final interpretive layer. This division of labor is consistent with the intended use of \textit{pArticleMap}: an auditable assistant for literature-grounded hypothesis generation, not a replacement for scientific judgment.

\section{Results \& Discussion}\label{results}

We evaluate \textit{pArticleMap} on four cue-conditioned retrospective bundles covering vaccines, antimicrobials, payload integration, and biosensing. Across these bundles and under the benchmark’s task-retention protocol, 195 task-level retained hypotheses~(winner ideas) were evaluated. The results below focus on observed realization behavior, winner-selection lift, and the extent to which internal scoring aligns with blinded human review.

\subsection{Observed retrospective realization performance}

Table~\ref{tab:retrospective-main} summarizes the task-retained hypotheses performance of \textit{pArticleMap}. Exact paper-level realization remained deliberately difficult: winner ideas achieved a gold-recovered rate of 10.8\%, recall@10 of 15.9\%, and mean reciprocal rank of 0.083. At the same time, the system was rarely completely off-target. Among the 195 task-level retained and evaluated ideas, 61.0\% were labeled \texttt{future neighborhood}, 27.7\% were labeled \texttt{historical confound}, and only a single winner idea (0.5\%) was labeled \texttt{not recovered}. In practice, this means that \textit{pArticleMap} often reached the correct forward-looking neighborhood even when it did not isolate the exact held-out paper.

This pooled pattern is important for interpreting the benchmark correctly. The benchmark is stricter than anecdotal rediscovery because it requires historical grounding and penalizes strong pre-cutoff analogues, but exact realization is still only one possible success mode. A \texttt{future neighbor only} outcome can still reflect a scientifically meaningful bridge idea that lands in the right later design space without matching the exact paper chosen as gold.

\begin{table*}[t]
\centering
\caption{Pooled retrospective realization metrics for retained winner ideas produced by \textit{pArticleMap} across the four completed cue-conditioned bundles.}
\label{tab:retrospective-main}
\small
\begin{threeparttable}
\begin{tabular}{lrrrrrrrrr}
\toprule
Cohort & Tasks & Gold rec. (\%) & Future neighborhood (\%) & Hist. conf. (\%) & Not rec. (\%) & R@1 (\%) & R@5 (\%) & R@10 (\%) & MRR \\
\midrule
\textit{pArticleMap} & 195 & 10.8 & 61.0 & 27.7 & 0.5 & 3.1 & 8.2 & 15.9 & 0.083 \\
\bottomrule
\end{tabular}
\begin{tablenotes}[flushleft]
\footnotesize
\item Metrics are computed after retaining the task-best hypothesis in each review packet.
\item Gold recovery is stricter than recall@10 because a task is labeled as a historical confound when a strong pre-cutoff analogue already exists, even if the gold future paper is retrieved.
\end{tablenotes}
\end{threeparttable}
\end{table*}

\subsection{Domain heterogeneity across cue-conditioned tasks}

\begin{figure}[t]
    \centering
    
    \includegraphics[
        width=\linewidth
    ]{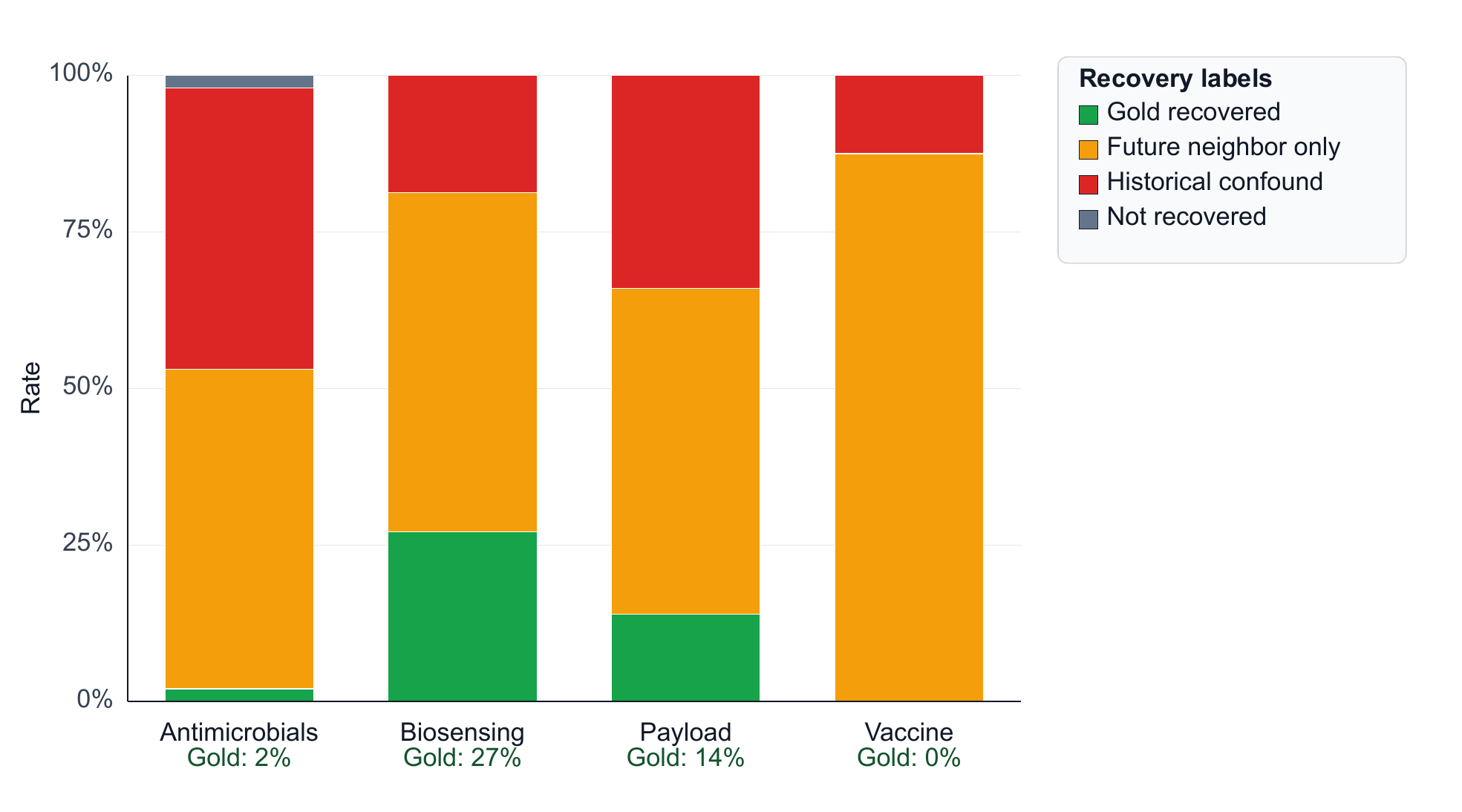}
    \caption{Winner-level recovery composition by domain for \textit{pArticleMap}. The dominant non-exact outcome is typically \texttt{future\_neighbor\_only}, with large cross-domain variation in exact realization and historical confounding.}
    \label{fig:winner-outcomes-by-domain}
\end{figure}

\begin{figure}[t]
    \centering
    \includegraphics[
        width=\linewidth
    ]{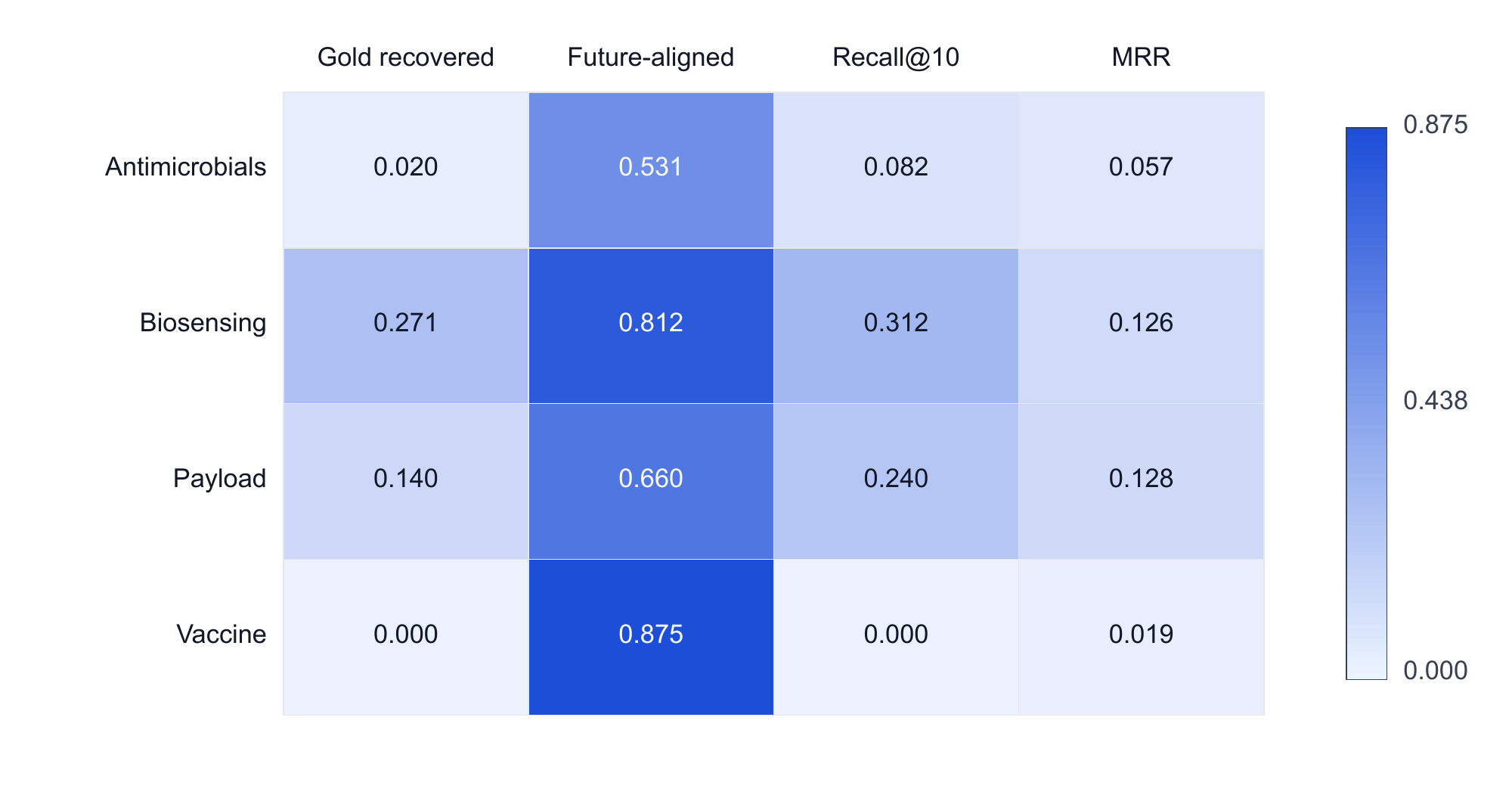}
    \caption{Winner-level realization metrics by domain for \textit{pArticleMap}. Biosensing and payload integration act as positive cases, whereas antimicrobials and vaccines expose confounding and exact-disambiguation difficulty.}
    \label{fig:winner-metric-heatmap}
\end{figure}

Realization behavior varied sharply by domain, as shown in Figures~\ref{fig:winner-outcomes-by-domain} and~\ref{fig:winner-metric-heatmap}. Biosensing gave the strongest exact-recovery signal, with 27.1\% gold recovery and 31.3\% recall@10, followed by payload integration with 14.0\% gold recovery and 24.0\% recall@10. Antimicrobials and vaccines were much harder. Antimicrobials produced only 2.0\% exact recovery and the highest historical-confound rate (44.9\%), while vaccines produced no exact gold recoveries despite 87.5\% \texttt{future\_neighbor\_only} outcomes.

This domain dependence is a substantive result rather than nuisance variance. Biosensing and payload integration behave like positive cases in which the cue and frontier structure are narrow enough to support both future-neighbor recovery and a nontrivial rate of exact realization. By contrast, antimicrobials and especially vaccines behave as stress tests: \textit{pArticleMap} still lands in the correct region of the future literature, but exact disambiguation is repeatedly blocked either by historically available analogues or by broad semantically similar future papers. In other words, failure is not monolithic. A high \texttt{future neighbor only} rate can still indicate historically grounded anticipation of the correct scientific neighborhood.

\subsection{Winner-selection lift in \textit{pArticleMap}}

\begin{figure}[t]
    \centering
    \includegraphics[
        width=\linewidth
    ]{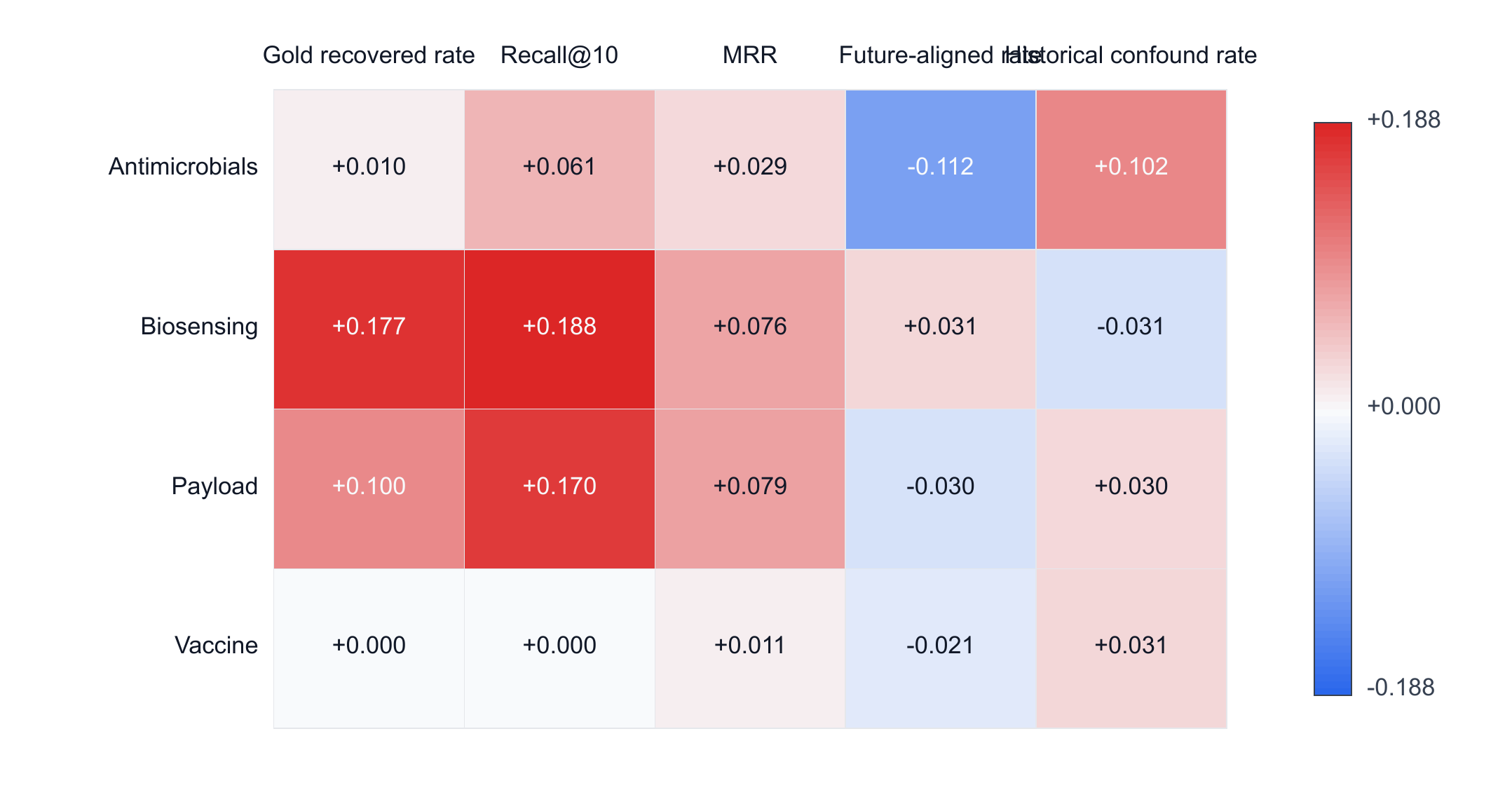}
    \caption{Winner-minus-nonwinner metric lift for \textit{pArticleMap}. Review-packet winner selection materially improves exact realization metrics, but does not reduce historical confounding.}
    \label{fig:winner-lift-heatmap}
\end{figure}

The winner-selection layer was useful even though it did not solve the full ranking problem. Relative to non-winner ideas, retaining the review-packet winner increased gold recovery from 3.6\% to 10.8\%, recall@10 from 5.4\% to 15.9\%, and MRR from 0.034 to 0.083. Figure~\ref{fig:winner-lift-heatmap} shows this lift across the main realization metrics. The ranking stage therefore concentrates the strongest future-aligned ideas in the current \textit{pArticleMap} output.

At the same time, winner selection did not eliminate historical overlap. The winner subset had a slightly higher historical-confound rate than the non-winner subset (27.7\% versus 24.4\%). This is an important qualitative finding: the current ranking path is useful for concentrating promising ideas, but it does not by itself separate genuinely frontier hypotheses from historically available near-neighbors. That remains one of the main unresolved ranking problems in the present system.

\subsection{Human evaluation and judge calibration}

Human assessment was available for 137 ideas that could be matched to agent-scored ideas across the four domains. With three reviewers per topic, the score-bearing workbooks contained 2,402 individual criterion scores; these were collapsed to 822 matched idea--criterion pairs by averaging the available reviewer scores for each idea and criterion. Pooled across all criteria, correspondence between human means and agent scores was positive but modest (Pearson $r=0.219$, Spearman $\rho=0.238$). Agreement was strongest for feasibility ($\rho=0.307$), followed by novelty ($\rho=0.212$) and evaluability ($\rho=0.156$). Importance and likely impact remained near zero, and plausibility was slightly negative ($\rho=-0.067$).

The calibration bias was also asymmetric. On average, \textit{pArticleMap}'s internal judge rated matched ideas 0.29 points higher than the human reviewers on the 5-point scale. The largest upward shifts were for importance, evaluability, and likely impact, whereas human reviewers rated plausibility higher than the agent and novelty remained almost perfectly aligned in the mean. Reviewer dispersion was itself informative: across-reviewer variation was largest for likely impact and importance, particularly in biosensing, indicating that the most strategic criteria were also the most subjective. Domain-level agreement remained modest, ranging from pooled Spearman $\rho=0.113$ in biosensing to $\rho=0.346$ in vaccines. Notably, the domain with the strongest exact realization signal, biosensing, was also the one with the weakest pooled human--agent agreement. Retrospective realization and expert-rated idea quality therefore probe different aspects of system behavior and should not be collapsed into a single scalar notion of idea quality.

\begin{table*}[b]
\centering
\caption{Pooled human-versus-agent calibration summary for the human-assessed \textit{pArticleMap} ideas, using three reviewers per topic.}
\label{tab:human-agent-calibration}
\small
\begin{threeparttable}
\begin{tabular}{lrrrr}
\toprule
Metric & Human mean & Agent mean & Spearman $\rho$ & Mean diff.\ (human $-$ agent) \\
\midrule
Importance & 2.619 & 3.496 & 0.066 & -0.871 \\
Novelty & 3.269 & 3.255 & 0.212 & 0.024 \\
Plausibility & 2.932 & 2.511 & -0.067 & 0.433 \\
Feasibility & 3.006 & 3.139 & 0.307 & -0.107 \\
Evaluability & 3.463 & 4.139 & 0.156 & -0.662 \\
Likely impact & 2.523 & 3.095 & 0.061 & -0.584 \\
\bottomrule
\end{tabular}
\begin{tablenotes}[flushleft]
\footnotesize
\item Human scores are mean reviewer scores on matched idea--metric pairs; when one reviewer did not provide a score for a specific criterion, the mean uses the available reviewer scores. Agent scores are the corresponding internal judge scores for the same matched ideas.
\end{tablenotes}
\end{threeparttable}
\end{table*}

\begin{rhoenv}[frametitle=Example Generated Hypotheses]
\textbf{Title.} \emph{Injectable supramolecular hydrogel depots that locally retain mRNA-LNPs and a Th1-biasing immunostimulant will improve durability of vaccine responses versus bolus LNP injection}. \\
\textbf{Short form.} The generated bridge hypothesis proposed embedding mRNA-LNPs and an added immunostimulant in a shear-thinning supramolecular hydrogel depot so that local antigen and adjuvant exposure would persist longer than with free LNP injection. It was grounded in evidence for sustained nanoparticle release, long-term local depots, and biomaterial-associated immune modulation, but it extrapolated those findings into a vaccine setting not directly represented in the retrieved pack. \\
\textbf{Human review.} Mean human score: 3.75/5 (importance 3.5, novelty 4.0, plausibility 4.0, feasibility 3.5, evaluability 4.0, likely impact 3.5). The audit still marked the bridge as under-supported and requested direct evidence for mRNA-LNP adjuvant inclusion and long-term efficacy. The full text appears in Appendix~\ref{appendix:worked-vaccine}.
\end{rhoenv}

\begin{figure}[hb]
    \centering
    
    \includegraphics[
        width=\linewidth
    ]{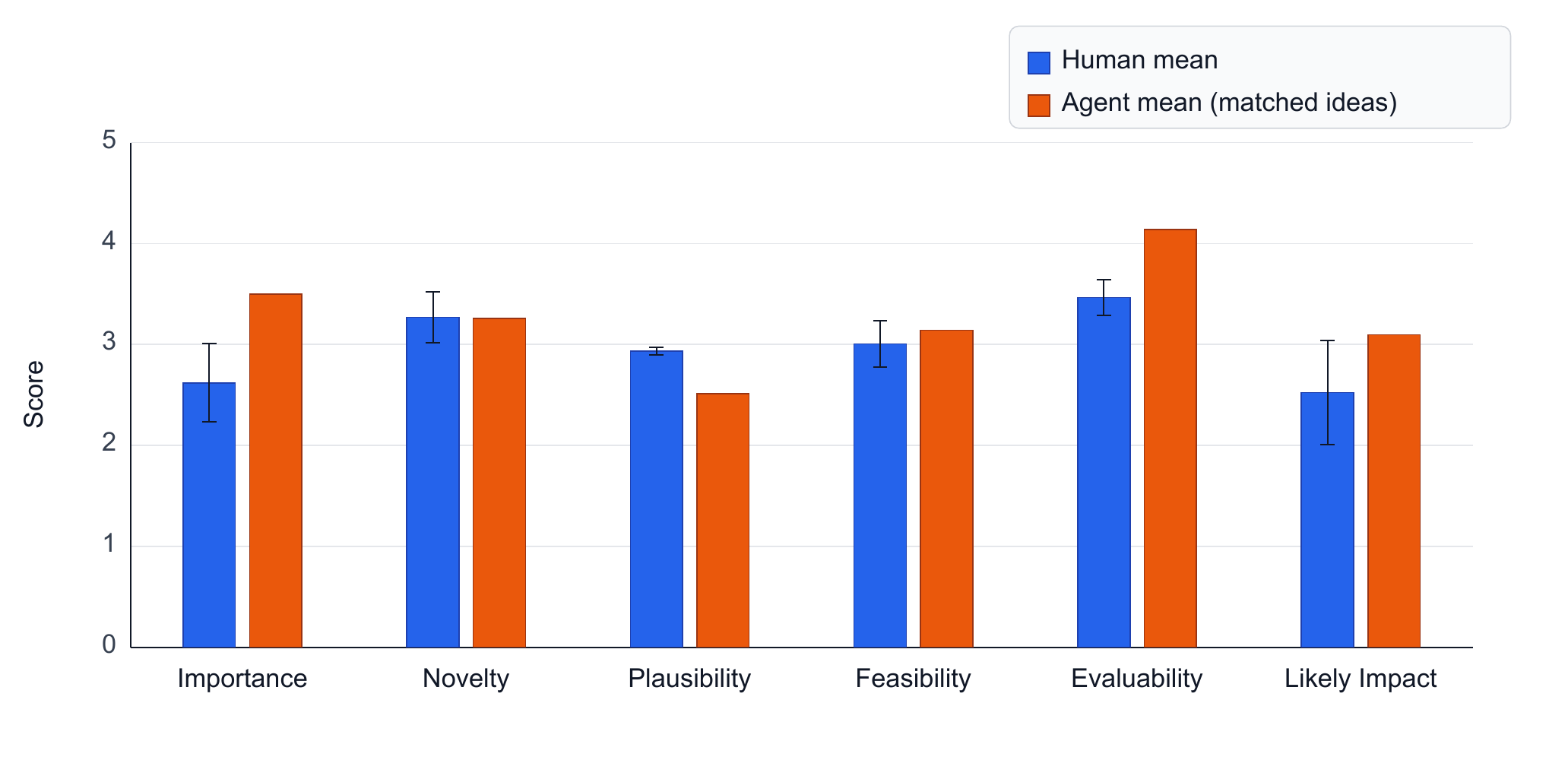}
    \caption{Bar graph with pooled human-versus-agent calibration summary for the \textit{pArticleMap} ideas. Error bars represent disagreement on the average score across the reviewers.}
    \label{fig:human-agent-eval}
\end{figure}

\subsection{Interpretation of \textit{pArticleMap}}

\newcolumntype{C}[1]{>{\centering\arraybackslash}p{#1}}
\newcolumntype{L}[1]{>{\raggedright\arraybackslash}p{#1}}
\begin{table*}[t]
    \caption{Mean human / agent evaluation scores (5-point scale) for the human-assessed hypotheses in each domain, using three reviewers per topic.}
    \label{tab:human-agent-domain-scores}
    \centering
    \small
    \begin{tabular}{
        L{0.1\textwidth}
        C{0.1\textwidth}
        C{0.1\textwidth}
        C{0.1\textwidth}
        C{0.1\textwidth}
        C{0.1\textwidth}
        C{0.1\textwidth}|
        C{0.1\textwidth}
    }
        \toprule
          & \multicolumn{6}{c}{Criteria} & \\
        \textbf{Domain} & Importance & Novelty & Plausibility & Feasibility & Evaluability & Likely impact  & Average \\
        \midrule
        Vaccines & 2.82 / 3.31 & 3.33 / 3.11 & 2.94 / 2.25 & 3.03 / 2.89 & 3.54 / 4.03 & 2.66 / 2.81 & 3.05 / 3.07 \\
        Antimicrobials & 3.02 / 3.35 & 3.24 / 3.07 & 2.98 / 2.70 & 3.19 / 3.74 & 3.40 / 4.30 & 2.84 / 3.02 & 3.11 / 3.36 \\
        Payload Int. & 2.65 / 4.00 & 3.39 / 3.79 & 2.81 / 2.90 & 2.67 / 2.79 & 3.36 / 4.26 & 2.60 / 3.63 & 2.91 / 3.56 \\
        Biosensing & 1.93 / 3.61 & 3.18 / 3.36 & 2.93 / 2.33 & 2.95 / 2.81 & 3.51 / 3.97 & 1.96 / 3.19 & 2.74 / 3.21 \\
        \midrule
        Overall & 2.62 / 3.50 & 3.27 / 3.26 & 2.93 / 2.51 & 3.01 / 3.14 & 3.46 / 4.14 & 2.52 / 3.10 & 2.97 / 3.27 \\
        \bottomrule
    \end{tabular}
\end{table*}
Taken together, these results position \textit{pArticleMap} as a conservative research-assistance system rather than an autonomous discovery engine. The strongest empirical claim is not that the system predicts the future literature verbatim. Instead, the evidence supports three narrower conclusions.

First, \textit{pArticleMap} is capable of historically grounded frontier targeting that often reaches the correct future neighborhood. This is the main significance of the large \texttt{future neighborhood} share and the low rate of total misses. Second, realization quality is strongly domain dependent. The system behaves best when the target frontier is narrow enough to support precise bridging and worst when historical analogues or broad future neighborhoods make exact disambiguation difficult. Third, internal groundedness scoring and retrospective realization do not replace expert review. The three-reviewer analysis strengthens the evidence that the current judge is useful as a support signal, especially for practical actionability, but not as a reliable surrogate for scientific judgment. The judge is directionally calibrated for some criteria, most clearly feasibility and novelty, yet it remains optimistic about strategic value and impact.

A further implication is that these retrospective results probably understate some aspects of prospective practical use while overstating the difficulty of exact benchmark matching. The four reported bundles were intentionally broad cue-conditioned stress tests: each cue carved out a coherent nanomedicine theme, but the resulting snapshots still contained wide design spaces and many semantically adjacent future papers. In a real prospective use case, an analyst would usually define a much narrower snapshot and cue, for example around a specific payload class, biological barrier, formulation constraint, assay system, or translational bottleneck. Such focus should make evidence packs more locally relevant and make generated hypotheses easier for experts to triage. At the same time, narrower prospective scoping is not simply an easier version of the retrospective benchmark. It may improve actionability and reduce irrelevant neighborhood matches, but it can also reduce serendipitous cross-domain bridging. The present results should therefore be interpreted as a coarse-grained validation of historically grounded frontier targeting, not as a direct estimate of performance in a focused prospective design session.

The human evaluation layer has a related temporal limitation. Although reviewers were blinded to the retrospective match labels and future retrievals, they could not be blinded to the scientific state of the field at the time of review. A hypothesis that would have been strategically important at the 2019 cutoff may appear less impactful after the relevant direction has become familiar, technically absorbed, or crowded by later work. Conversely, reviewers may also judge plausibility through knowledge that was unavailable at the cutoff date. This hindsight effect is most likely to affect importance and likely impact, precisely the criteria where reviewer dispersion was largest and human--agent agreement was weakest. For that reason, the human scores should be read as present-day expert calibration of generated ideas rather than as a pure reconstruction of their cutoff-date value.

These findings also clarify the current claim boundary of this paper. The present manuscript evaluates \textit{pArticleMap} directly and reports observed realization, ranking, and calibration behavior from the completed retrospective bundles. It does not make stronger claims about superiority over all simpler literature-mining baselines, because a fully integrated comparison across all alternative generation methods is not presented here. The resulting framing is intentionally conservative: \textit{pArticleMap} is valuable because it surfaces sparse article-level frontiers, assembles auditable evidence around them, and produces testable bridge hypotheses that can later be inspected against future literature and human judgment.

\section{Limitations \& Future Work}\label{limitations}

\subsection{Current limitations}

\textit{pArticleMap} should be interpreted as a literature-grounded hypothesis-generation assistant, not as a complete discovery engine. Its first major limitation is corpus scope. The dataset is derived from a PubMed-centered query emphasizing nanoparticles and in vivo nanomedicine language. That choice yields a large and useful corpus for preclinical nanomedicine, but it inevitably omits adjacent evidence channels such as patents, clinical-trial records, conference proceedings, citation networks, negative results, and much of the broader biomaterials and drug-delivery ecosystem. Even within PubMed, the canonical representation is title plus abstract rather than full text, which constrains how much mechanistic detail can be retrieved and compared reproducibly.

The corpus is also linguistically and epistemically narrow. Most records are English-language, which improves consistency but introduces selection bias. In addition, the source query privileges published preclinical practice over translational readiness. As a result, the literature map is better at revealing what has been discussed in experimental nanomedicine than at revealing what is clinically mature, manufacturable, or regulatorily realistic.

Methodologically, low-density regions in embedding space remain only a proxy for scientific opportunity. Sparse neighborhoods may indeed correspond to overlooked conceptual bridges, but they may also reflect naming variation, incomplete indexing, unstable terminology, or artifacts of the selected encoder, PCA projection, neighborhood sizes, and clustering method. The present implementation of \textit{pArticleMap} therefore helps prioritize frontier candidates; it does not prove that a sparse region is important or even meaningful in a causal scientific sense.

The agentic layer introduces its own limitations. Evidence packs, citation requirements, and audit-driven retrieval patching reduce unsupported generation, but they do not eliminate it. The quality of the generated hypotheses still depends on retrieval coverage, evidence-pack composition, prompt design, and the model's ability to summarize heterogeneous abstracts faithfully. A citation-backed statement can still be weak, overly incremental, or misleading if the retrieved evidence is incomplete, imbalanced, or ambiguous.

Evaluation is likewise a proxy problem rather than a definitive scientific test. The retrospective benchmark measures whether hypotheses generated from historical evidence retrieve later literature and land in the correct future neighborhood under a leakage-controlled historical split. This is methodologically stronger than anecdotal rediscovery, but it still does not establish mechanistic truth, experimental feasibility, or clinical value. The matching layer depends on heuristic fingerprints, reranking thresholds, and approximate semantic overlap. A future match can therefore reflect broad topical similarity rather than a genuinely prescient mechanistic insight, while a strong idea may be undercounted if it is expressed differently from the later literature.

\subsection{Future work}

The most immediate extension is to broaden the evidence substrate. Future versions of \textit{pArticleMap} should integrate full-text open-access papers where licensing permits, citation graphs, biomedical ontologies such as MeSH and UMLS, and domain-specific nanomaterial resources such as eNanoMapper and related nanosafety repositories. Doing so would reduce the current dependence on abstract-level summaries and should improve both evidence retrieval and downstream experimental blueprints.

On the representation side, \textit{pArticleMap} would benefit from explicit robustness and ablation studies. Important comparisons include encoder choice, PCA versus original embedding-space analysis, neighborhood scales, clustering method, evidence-pack composition, and audit-loop depth. Stability analysis over bootstrap resampling or rolling temporal snapshots would help separate persistent structural gaps from transient sparsity artifacts.

The evaluation protocol should also move beyond a single historical split. Rolling-origin benchmarks across multiple cutoffs, combined with blinded expert review, would provide a more defensible assessment of novelty, plausibility, grounding quality, and actionability. Ultimately, the strongest validation would be prospective rather than retrospective: generating hypotheses from a frozen snapshot, selecting a small number for expert triage, and following them through real experimental or translational decision processes.

From a nanomedicine perspective, future work should move closer to the constraints that matter in practice. Generated hypotheses and blueprints should represent manufacturability, safety, dosing route, biomarker strategy, regulatory terminology, and model-system relevance more explicitly. That shift would make the system more useful for translational reasoning rather than only for literature novelty.

Finally, the longer-term opportunity is not full autonomy but stronger human-AI collaboration. A mature version of \textit{pArticleMap} should allow domain experts to inspect uncertainty, challenge evidence selection, inject structured counter-evidence, compare alternative bridge hypotheses, and accumulate review signals over time. In that setting, the system becomes less a generator of polished text and more a reproducible interface between large-scale literature analysis and expert scientific judgment.

\section{Conclusion}\label{conclusion}

We present \textit{pArticleMap}, an evidence-grounded literature-mapping and hypothesis-generation system for nanomedicine that combines article-level representation learning, graph-based frontier detection, snapshot-backed evidence retrieval, and an audited multi-step generation workflow. The central contribution is not a single language-model prompt, but the integration of target selection, evidence construction, explanation, audit, retrieval patching, ideation, scoring, and blueprint generation into one reproducible pipeline with explicit provenance.

Just as importantly, we evaluate \textit{pArticleMap} conservatively. Rather than treating plausible output text as evidence of discovery, we pair the system with a retrospective realization benchmark and a blinded human-review layer. Across four cue-conditioned retrospective bundles, \textit{pArticleMap} frequently reached the correct future neighborhood even when exact paper-level realization remained difficult, and the human-review results showed that internal idea scoring captures only part of the practical-actionability signal. These findings support \textit{pArticleMap} as a useful discovery-support tool, while also making clear that retrieval-based realization and internal scoring do not eliminate the need for expert scientific judgment.

The resulting picture is deliberately practical. \textit{pArticleMap} should be understood as a grounded nanomedicine research assistant that helps users surface sparse interfaces in a large literature, inspect the evidence around those interfaces, and formulate more auditable and testable research directions. It does not establish mechanistic truth, translational value, or clinical readiness on its own. With broader evidence sources, stronger robustness studies, and prospective expert-led validation, \textit{pArticleMap} could evolve into a durable interface between large-scale literature mapping and nanomedicine research design.
\newpage


\printbibliography
\newpage
\appendix
\section{Corpus-Construction Flow}\label{appendix:corpus-flow}

Section~\ref{methods} gives the formal definition of corpus assembly, representation, graph construction, clustering, and gap extraction. This section provides the corresponding visual walk-through. The numbered figures in this section use the vaccine-focused slice as a representative example, but the same sequence is used for the full PubMed-derived corpus and for the other cue-conditioned semantic subsets.

The purpose of this appendix is therefore complementary rather than repetitive. In the main methods section, the \textit{pArticleMap} workflow is defined mathematically and operationally; here, the same workflow is shown as an end-to-end progression from a broad corpus landscape, to a semantically filtered subcorpus, to a local similarity graph, to operational communities, and finally to stored sparse-region targets that drive downstream evidence-pack construction.

\subsection{From the full corpus to a domain-focused slice}

\begin{figure}[ht]
    \centering
    \begin{minipage}[t]{0.49\textwidth}
        \centering
        \includegraphics[scale=0.05]{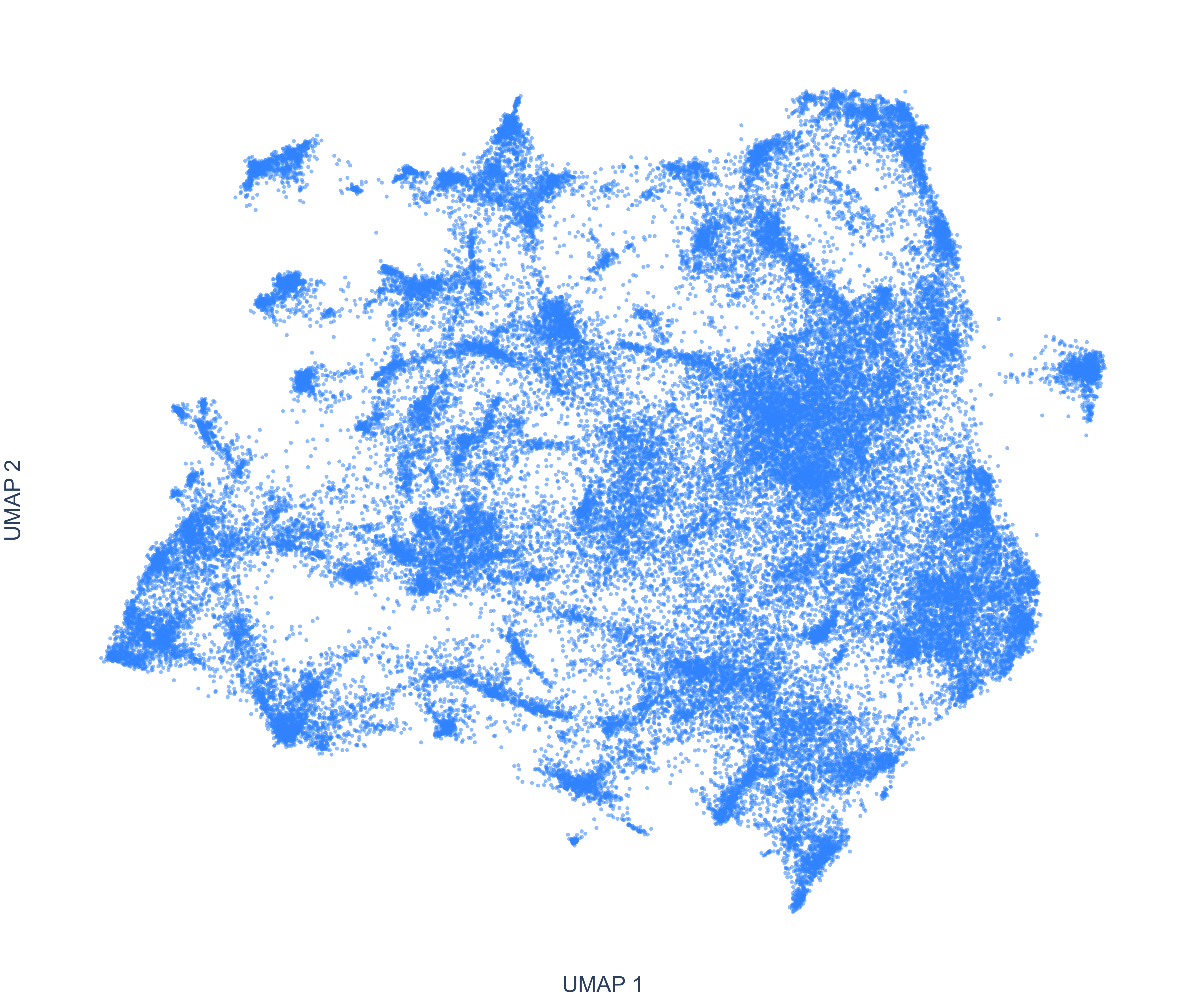}
    \end{minipage}\hfill
    \begin{minipage}[t]{0.49\textwidth}
        \centering
        \includegraphics[scale=0.05]{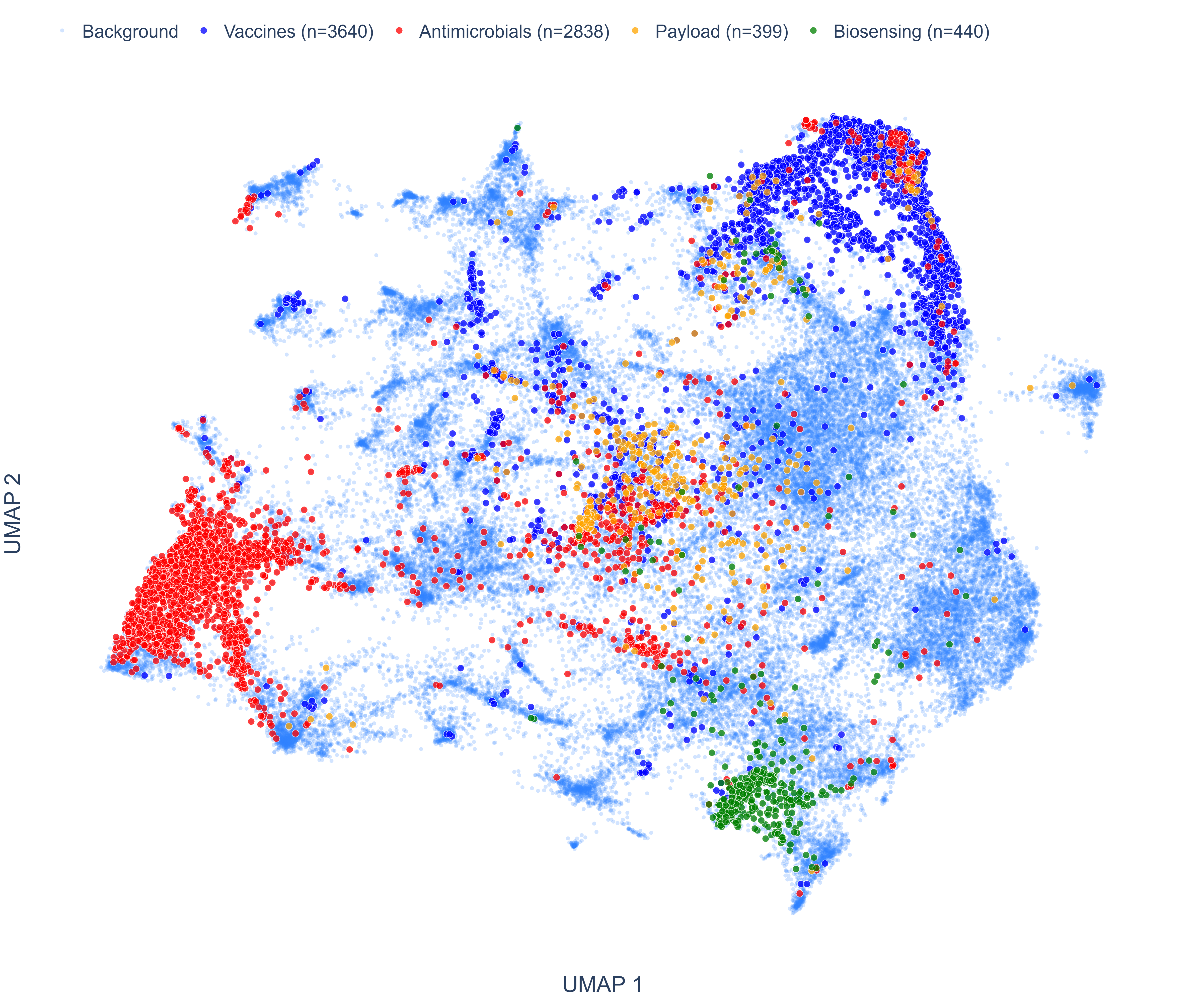}
    \end{minipage}
    \caption{Corpus overview and semantic slicing. Left: global UMAP view of the full nanomedicine corpus, used as an exploratory map rather than as an analysis space. Right: overlay of semantic filters showing how domain-specific queries occupy coherent but partially overlapping regions of the broader literature landscape.}
    \label{fig:corpus-overview}
\end{figure}

The first stage is corpus inspection and semantic scoping. Figure~\ref{fig:corpus-overview} shows the global corpus projection together with the overlay of semantic filters used to carve out domain-specific subcorpora. As emphasized in Section~\ref{methods}, UMAP is used here only for visualization. The actual downstream analysis is still performed in the embedding or PCA-reduced analysis space, not in the plotted coordinates.

\begin{figure}[ht]
    \centering
    \begin{minipage}[t]{0.49\textwidth}
        \centering
        \includegraphics[scale=0.05]{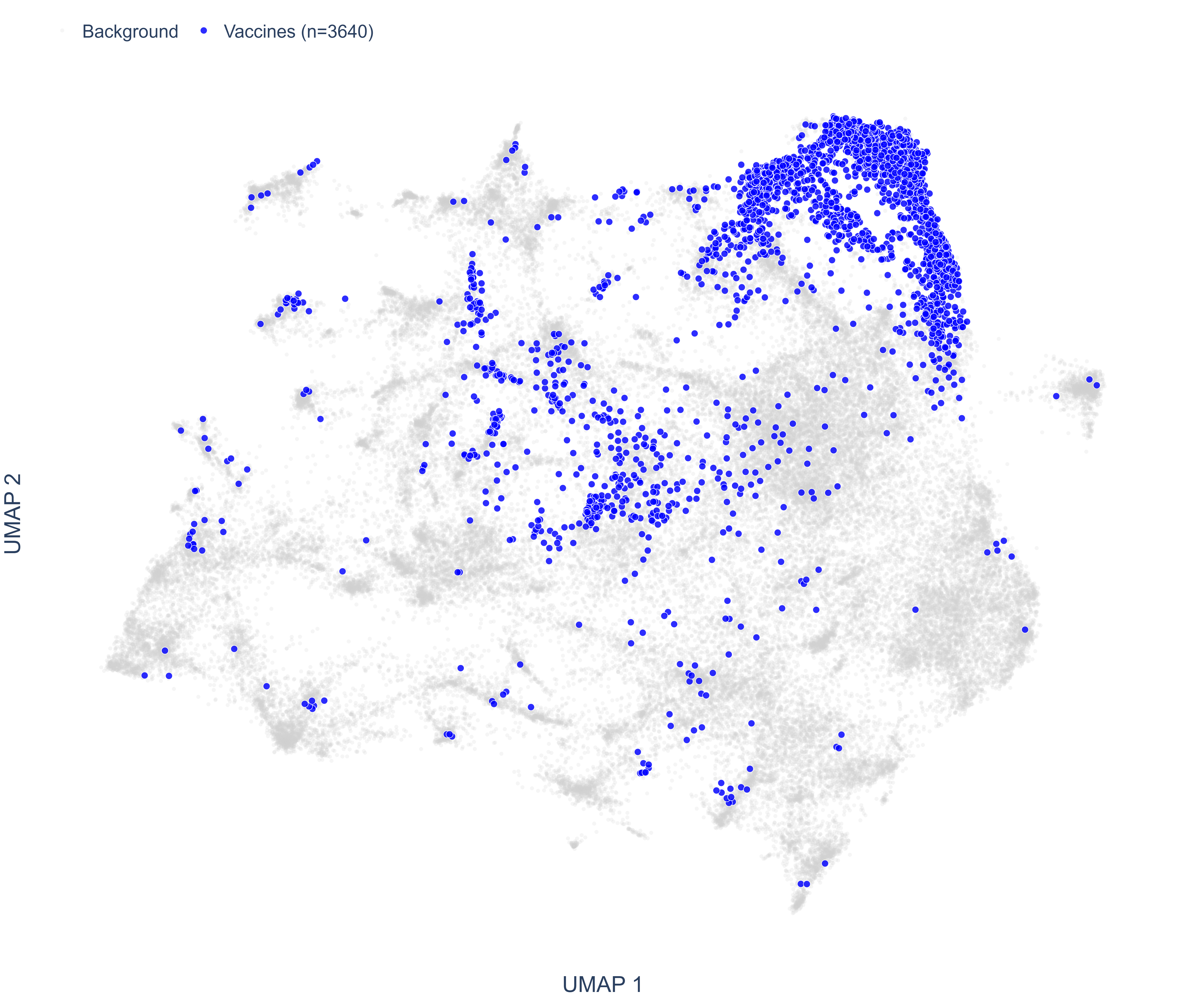}
    \end{minipage}\hfill
    \begin{minipage}[t]{0.49\textwidth}
        \centering
        \includegraphics[
            width=\linewidth
        ]{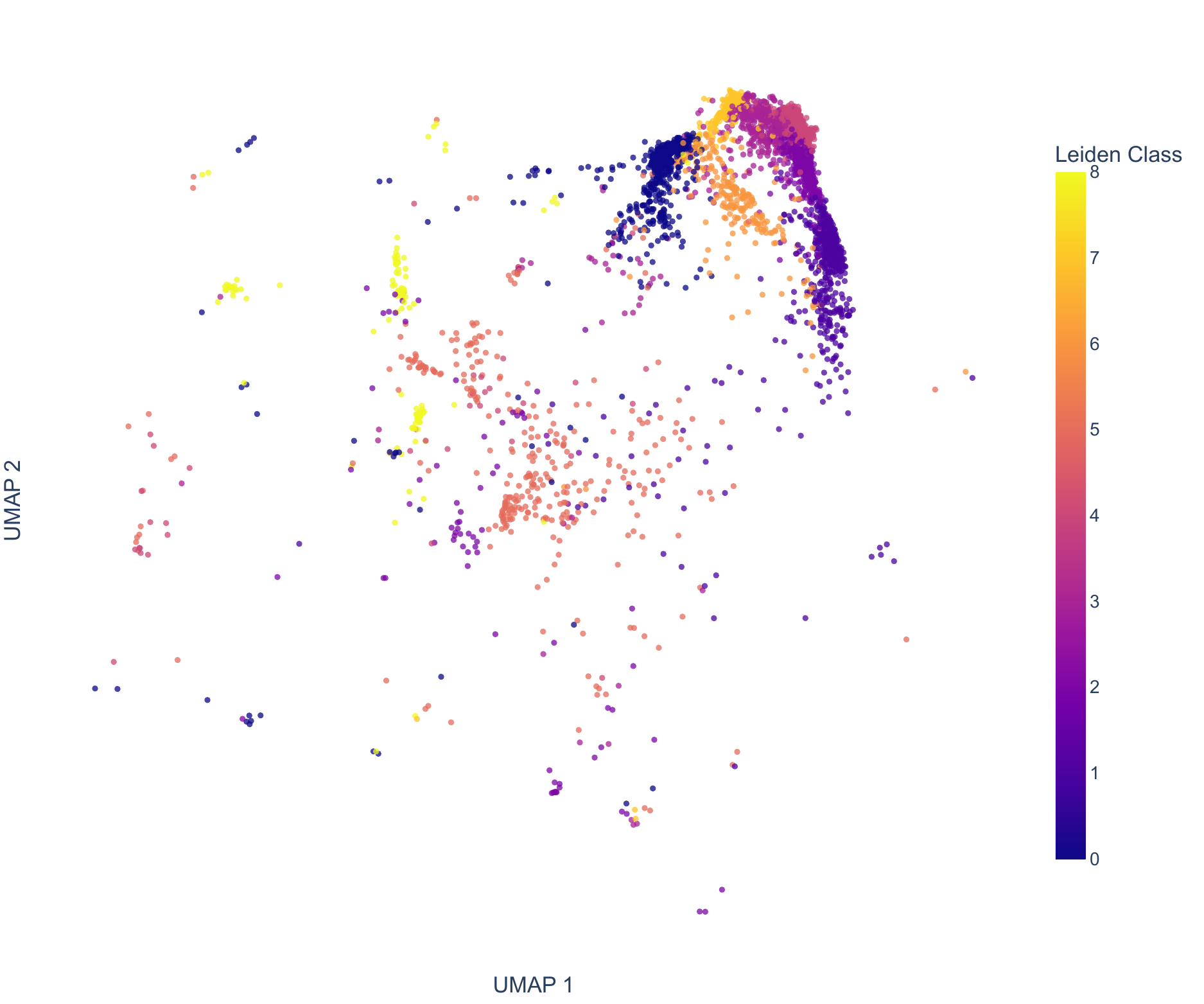}
    \end{minipage}
    \caption{Semantic-thresholded subcorpus and operational communities for the vaccines example. Top: papers retained after semantic filtering and thresholding. Bottom: Leiden communities computed on the similarity graph, providing the operational clustering propagated into the stored snapshot. }
    \label{fig:corpus-filter-graph}
\end{figure}

After an analyst defines one semantic direction through text-conditioned embedding similarity, \textit{pArticleMap} thresholds that slice to produce a retained subcorpus for detailed analysis. Figure~\ref{fig:corpus-filter-graph} shows the vaccines example: first the thresholded vaccine-relevant papers, then the similarity graph built over that retained subset. This step connects the broad corpus-construction logic in Section~\ref{methods} to the practical object that is later clustered and scored.

\subsection{Similarity graph to operational literature communities}

\begin{figure}[t]
    \centering
    \begin{minipage}[t]{0.49\textwidth}
        \centering
        \includegraphics[
        width=\linewidth
    ]{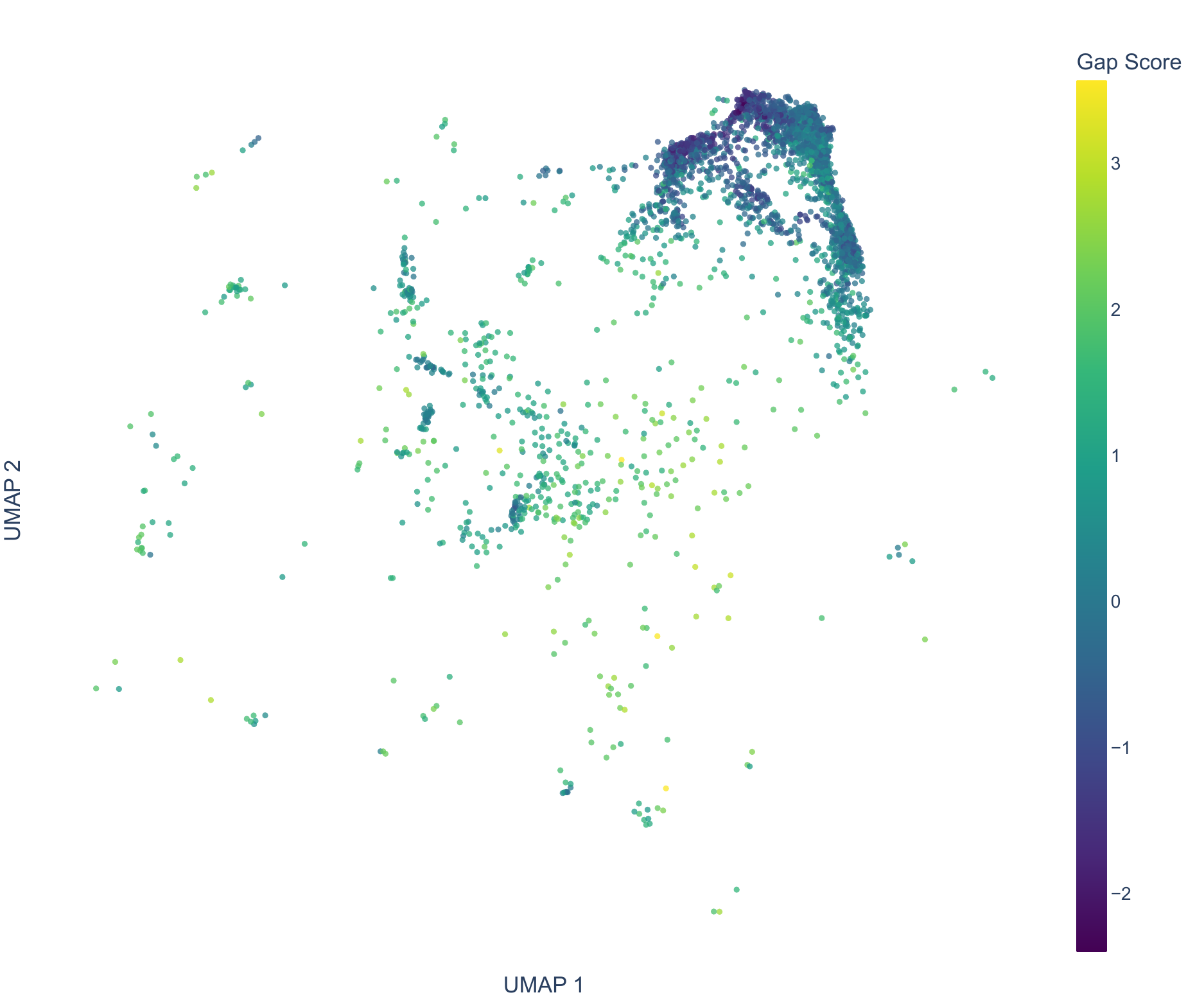}
    \end{minipage}
    \caption{Paper-level gap scores over the same slice, highlighting sparse frontier papers at the interface of denser literature regions.}
    \label{fig:corpus-cluster-gapscore}
\end{figure}

\begin{figure*}[t]
    \centering
    \begin{minipage}[t]{0.49\textwidth}
        \centering
        \includegraphics[
        width=\linewidth
    ]{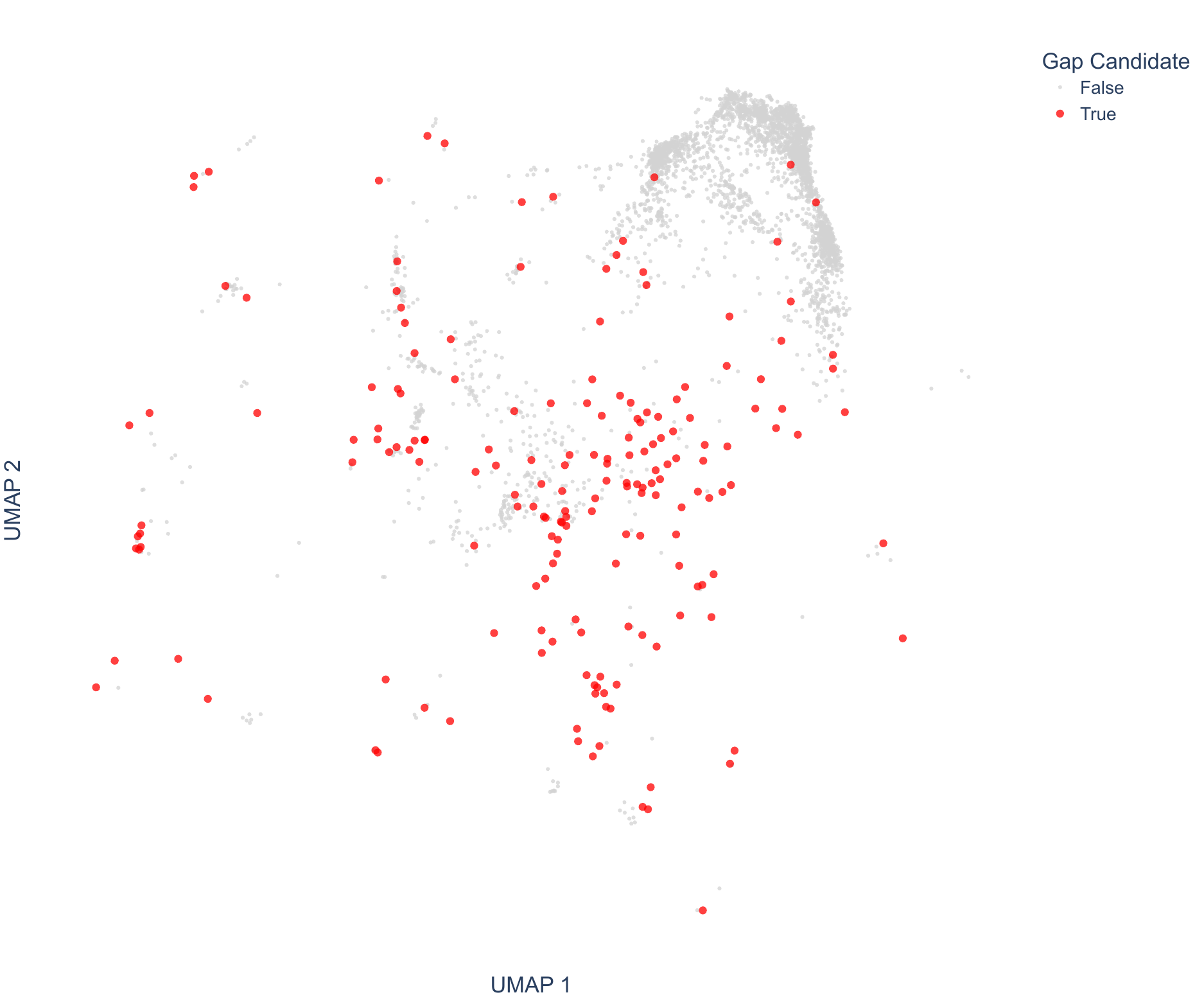}
    \end{minipage}\hfill
    \begin{minipage}[t]{0.49\textwidth}
        \centering
        \includegraphics[
        width=\linewidth
    ]{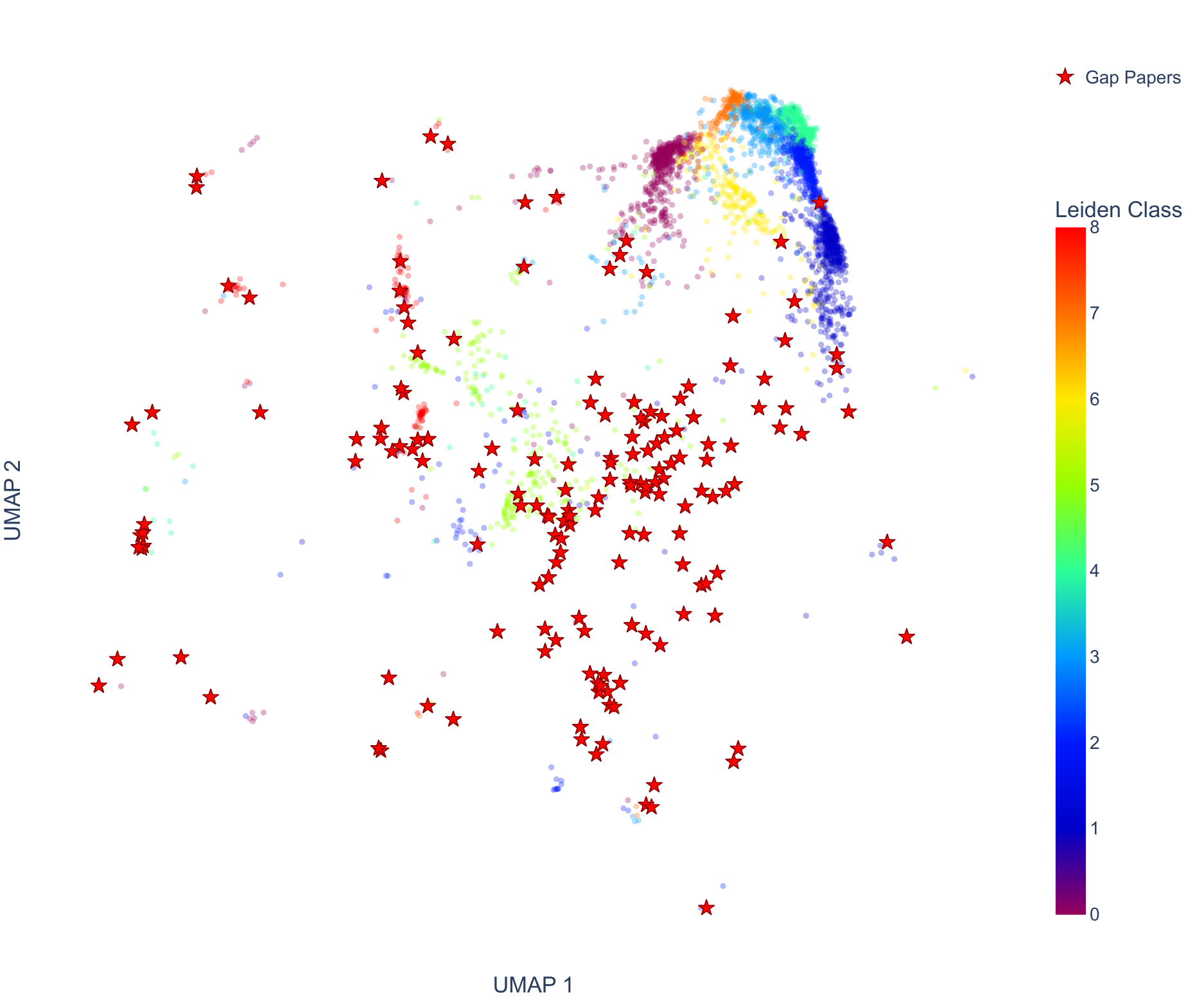}
    \end{minipage}
    \caption{Gap-region extraction and target surfacing. Left: retained connected components among the high-gap-score papers after thresholding and minimum-size filtering. Right: the same gap regions overlaid on Leiden communities, showing how sparse regions induce both direct gap targets and cluster-pair bridge targets for downstream evidence-pack construction.}
    \label{fig:corpus-gap-targets}
\end{figure*}

Once the retained slice has been converted into a graph, \textit{pArticleMap} computes one operational clustering and one multi-scale sparsity score per paper. Figure~\ref{fig:corpus-cluster-gapscore} visualizes the multi-scale sparsity score output, which identifies papers that remain locally isolated across several neighborhood scales rather than under a single arbitrary choice of $k$.

\subsection{From sparse papers to published bridge targets}

The final analysis stage turns a continuous sparsity field into discrete retrieval targets. Figure~\ref{fig:corpus-gap-targets} shows first the retained connected components among the high-gap-score papers, and then the same components overlaid on the Leiden partition. This pairing is important: the components define explicit gap targets, while the communities they touch define the cluster-pair bridge targets later used by the retrieval and agentic-generation stack.

Taken together, Figures~\ref{fig:corpus-overview}--\ref{fig:corpus-gap-targets} show the full corpus-creation and analysis flow used by \textit{pArticleMap}. The process begins with a broad PubMed-derived literature collection, narrows to a semantically meaningful analysis slice, imposes a local similarity graph, derives an operational community structure, scores local sparsity, and finally freezes the resulting papers, clusters, gap memberships, and metadata into a reusable snapshot. That snapshot is the object consumed later by the evidence-pack builder, orchestrator, and retrospective evaluation pipeline.
\newpage
\section{Agent Orchestration Workflow}\label{appendix:agent-workflow}

\begin{figure*}[ht!]
    \centering
    \includegraphics[width=2.0\columnwidth]{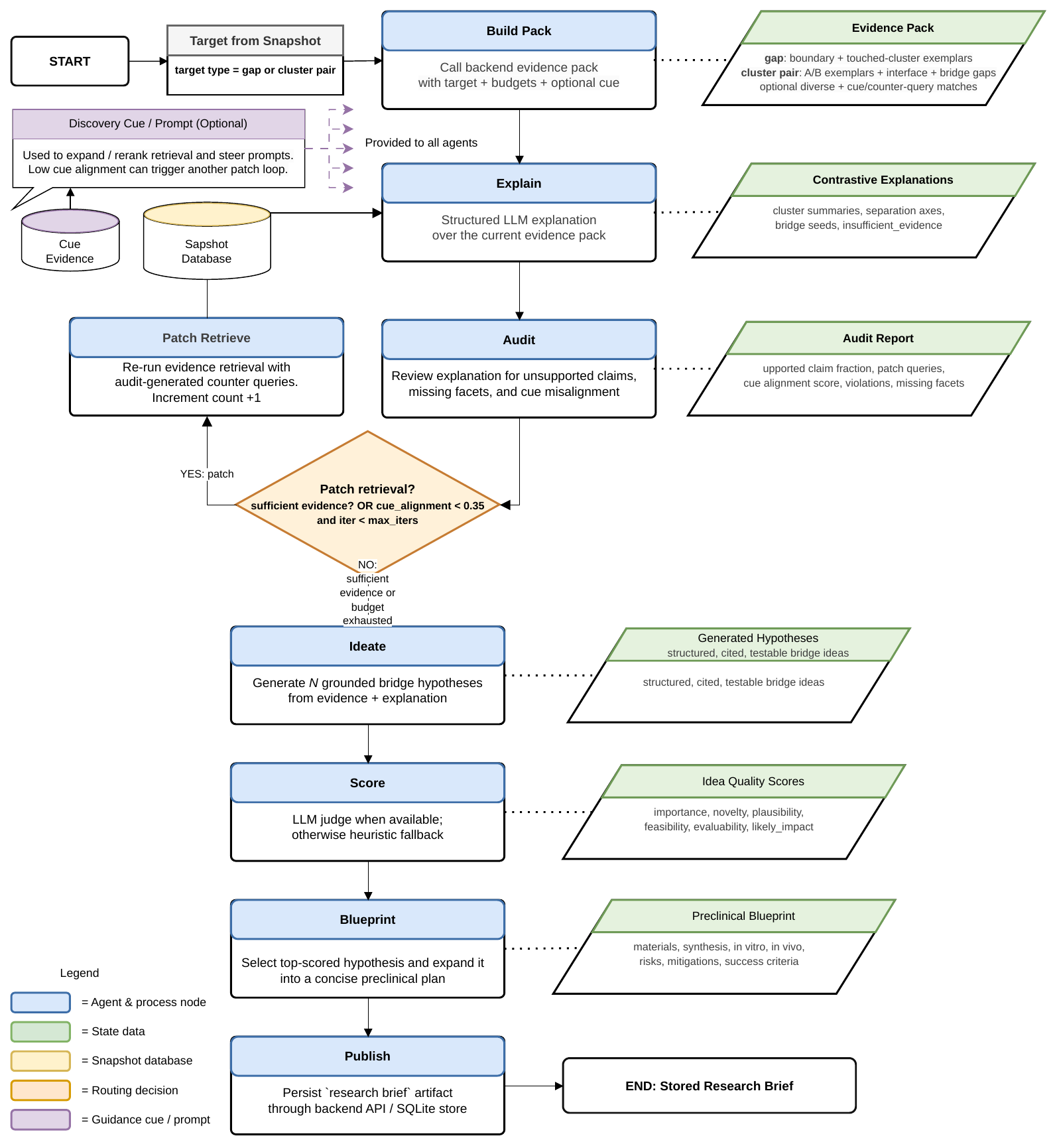}
    \caption{Implemented LangGraph orchestration path for \textit{pArticleMap}. The system builds a target-conditioned evidence pack, explains the frontier, audits the explanation, optionally patches retrieval, generates multiple grounded hypotheses, scores them, and emits a preclinical blueprint only for the top-scored idea.}
    \label{fig:langgraph-workflow}
\end{figure*}

Figure~\ref{fig:langgraph-workflow} shows the concrete execution path implemented by \textit{pArticleMap}. The runtime does not pass directly from retrieval to generation. Instead, it first assembles a target-conditioned evidence pack, produces a structured explanation of the frontier, audits that explanation for unsupported claims and cue misalignment, optionally repairs retrieval with auditor-proposed patch queries, and only then emits and scores bridge hypotheses before selecting one for blueprint generation.

This distinction matters for scientific use. The \texttt{research brief} stores the evidence pack, explanation payload, audit payload, hypothesis set, idea scores, blueprint, and trace metadata together in one auditable object. In the bundled antimicrobial export used for the examples below, the stored briefs all had a \texttt{iterations = 2} budget, meaning that each completed run executed one explanation--audit loop, one patch-retrieval pass, and a second explanation pass before ideation.

\section{Expected retrospective benchmark outcomes}\label{appendix:expected_outcome}

The central empirical hypothesis is that the full orchestrator should outperform simpler controls on future-paper recovery because it does more than generate from a retrieved pack once. It explains the frontier target, audits the sufficiency of evidence, patches retrieval when needed, and only then generates and scores bridge hypotheses. If that additional structure is useful, the expected qualitative ranking is the one summarized in Table~\ref{tab:expected_methods}.

\begin{table*}[t]
    \caption{Primary retrospective result structure and expected qualitative outcome pattern.}
    \label{tab:expected_methods}
    \centering
    \small
    \begin{tabular}{
        >{\raggedright\arraybackslash}p{0.16\textwidth}
        >{\raggedright\arraybackslash}p{0.15\textwidth}
        >{\raggedright\arraybackslash}p{0.16\textwidth}
        >{\raggedright\arraybackslash}p{0.15\textwidth}
        >{\raggedright\arraybackslash}p{0.28\textwidth}
    }
        \toprule
        \textbf{Method} & \textbf{Expected Recovery} & \textbf{Expected Confound Rate} & \textbf{Expected Cue Response} & \textbf{Expected Interpretation} \\
        \midrule
        orchestrator & Highest among grounded methods & Low to moderate & High & The explain-audit-patch loop should improve evidence sufficiency and produce hypotheses that are both more target-specific and more historically grounded. \\
        \midrule
        single shot LLM & Moderate & Moderate & Moderate & The evidence pack provides grounding, but the absence of iterative auditing should increase underspecified or weakly supported hypotheses. \\
        \midrule
        retrieval summary & Moderate & Moderate & Moderate to high & Pack summarization can improve compression and readability, but may also discard boundary details needed for sharper bridge hypotheses. \\
        \midrule
        heuristic bridge & Low & Low & Low & This baseline should remain useful as a non-agentic sanity check, but it is not expected to recover many exact future papers. \\
        \midrule
        pack query (baseline) & Low to moderate for nearby neighbors, low for exact recovery & Low & Low & Deterministic query generation should retrieve related literature, but it is not expected to compete with audited hypothesis generation on exact future-paper matching. \\
        \midrule
        random target control & Lowest & Variable & Lowest & Poor performance here would confirm that target assignment matters and that recovery is not explained by generic nanomedicine language alone. \\
        \bottomrule
    \end{tabular}
    \tabletext{Recovery refers primarily to \texttt{gold\_recall@k}, \texttt{gold\_recovered\_rate}, and \texttt{gold\_MRR}. Confound rate refers to \texttt{historical\_confound\_rate}.}
\end{table*}

The most important comparison is therefore not between the orchestrator and an unrealistically weak strawman, but between the full audited workflow and simpler retrieval-grounded alternatives that operate on the same historical targets. If the full workflow does not improve recovery over \texttt{single\_shot\_llm} or \texttt{retrieval\_summary\_direct}, then the additional orchestration complexity would be difficult to justify. Conversely, if the orchestrator improves future-paper recovery while keeping \texttt{historical\_confound\_rate} low, that would support the claim that iterative evidence repair is materially useful for scientific ideation.

Even under a successful outcome pattern, the benchmark should remain difficult. Exact \texttt{gold\_recovered} matches are stringent because a generated hypothesis must be close enough to a later paper to retrieve it near the top of the future corpus while also avoiding strong historical matches. For that reason, we expect many scientifically useful cases to fall into \texttt{future neighborhood} rather than \texttt{gold\_recovered}. In this setting, high-quality results should be interpreted as evidence of frontier alignment and historically grounded anticipation, not as proof that the system predicts the future literature verbatim. 

\subsection{Expected cue-conditioned domain behavior}

The four cue-conditioned corpora described in Section~\ref{retro_validation} probe a different question from the unconstrained benchmark: whether the same grounded workflow can be steered toward specific nanomedicine design problems without losing historical discipline. These tasks were chosen to span immunological design, antimicrobial materials, multifunctional payload engineering, and nucleic-acid biosensing. The expected qualitative behavior is summarized in Table~\ref{tab:expected_domains}.

\begin{table*}[t]
    \caption{Expected qualitative outcome pattern for the four cue-conditioned benchmark domains.}
    \label{tab:expected_domains}
    \centering
    \small
    \begin{tabular}{
        >{\raggedright\arraybackslash}p{0.13\textwidth}
        >{\raggedright\arraybackslash}p{0.27\textwidth}
        >{\raggedright\arraybackslash}p{0.24\textwidth}
        >{\raggedright\arraybackslash}p{0.24\textwidth}
    }
        \toprule
        \textbf{Domain} & \textbf{Scientific Focus} & \textbf{Expected Strongest Gain} & \textbf{Expected Main Challenge} \\
        \midrule
        Vaccines & Adjuvant design for durable immunity in mRNA LNP vaccination & Improved cue-weighted retrieval and stronger separation between delivery chemistry and immunological mechanism in the generated hypotheses & The surrounding literature is broad, so semantically related but non-vaccine immunology papers may compete strongly during retrieval and matching. \\
        Antimicrobials & Surface or coating properties that help inorganic nanoparticles overcome biofilms & Better targeting of materials-surface-biofilm interfaces that are often fragmented across materials and infection literatures & Broad terminology around coatings, infection, and antibiofilm activity may increase neighbor-only matches rather than exact gold recovery. \\
        Payload integration & Co-delivery of proteins and nucleic acids within one nanoparticle platform & Strongest expected cue benefit because the task explicitly requires bridging partially separated payload literatures & The matcher may confound truly multifunctional systems with generic co-delivery papers unless mechanistic detail is preserved in the fingerprint. \\
        Biosensing & Nanoparticle systems for simultaneous DNA and RNA detection & Higher cue alignment and more interpretable cluster-pair targets connecting sensing and nucleic-acid detection literatures & Retrieval may drift toward general molecular diagnostics unless the cue maintains dual-target specificity. \\
        \bottomrule
    \end{tabular}
\end{table*}

The expected benefit of cue steering is strongest in the cue-weighted metrics and in expert-rated relevance, not necessarily in unconditional future-paper recovery. This distinction matters. A discovery cue narrows the search toward a user-specified scientific direction, which should improve topical precision and actionability, but it can also reduce breadth and therefore make exact future-paper recovery harder in some settings. The scientifically useful outcome is therefore not simply ``cue beats no cue'' on every metric; it is that cue-active runs become more directionally coherent without collapsing into unsupported speculation.

\newpage
\section{Example from Antimicrobial Retrospective Run}\label{appendix:worked-antimicrobial}

To make the workflow concrete, we inspected the antimicrobial retrospective export used for the results in this project. This run used a historical cutoff date of December 31, 2019, a future evaluation window from January 1, 2020 to January 1, 2026, and the discovery cue: \emph{``What characteristics should a coating for inorganic nanoparticles have to overcome biofilms?''} The examples below were reconstructed jointly from the stored SQLite \texttt{research brief} artifacts, the assessment bundle, and the reviewer packet.

These examples are intentionally not cherry-picked as ``perfect'' successes. Instead, they illustrate \textit{pArticleMap}'s characteristic behavior in a difficult cue-conditioned domain: evidence packs successfully bridge delivery, materials, and imaging literatures; the audit stage remains conservative about unsupported biofilm-specific extrapolation; ideation produces several mechanistically distinct bridge candidates; and the blueprint stage expands only the internally top-scored idea, which is not always the same hypothesis later retained by retrospective future-paper matching.

\begin{table*}[t]
    \caption{Representative appendix examples from the antimicrobial retrospective export. ``Reviewer-retained idea'' refers to the hypothesis kept in the review packet for the downstream future-paper task, whereas ``blueprinted idea'' refers to the top internally scored hypothesis expanded by the runtime blueprint agent from the same trace.}
    \label{tab:appendix-antimicrobial-examples}
    \centering
    \small
    \begin{tabular}{
        >{\centering\arraybackslash}p{0.02\textwidth}
        >{\raggedright\arraybackslash}p{0.10\textwidth}
        >{\raggedright\arraybackslash}p{0.18\textwidth}
        >{\raggedright\arraybackslash}p{0.21\textwidth}
        >{\raggedright\arraybackslash}p{0.21\textwidth}
        >{\raggedright\arraybackslash}p{0.14\textwidth}
    }
        \toprule
        Ex. & Target & Explain-stage contrast & Audit diagnosis & Reviewer-retained idea / blueprinted idea & Retrospective outcome \\
        \midrule
        A &
        \texttt{cluster pair 0-3} \newline 64-paper pack &
        Responsive delivery and ROS/redox-triggered release versus multifunctional inorganic photothermal and imaging platforms with biocompatible surface engineering &
        High local support (\(\approx 0.96\)) for the frontier explanation, but explicit missing facets on EPS penetration, biofilm-compatible coating chemistry, trigger validity in biofilms, and safety &
        Retained: pH-sheddable, charge-reversible coating for transport then adhesion. \newline Blueprinted: polydopamine-coated photothermal inorganic nanoparticle with ROS/pH/NIR-triggered release &
        \texttt{future neighborhood}; nearest future match was a pH-triggered silver nanoparticle antibacterial therapy paper \\
        \midrule
        B &
        \texttt{gap 0} \newline 69-paper pack &
        Stealth and anti-clearance coatings versus matrix-interacting, targeted, or antimicrobial surface modifications &
        Weaker support (\(\approx 0.79\)); the auditor specifically noted lack of direct mature-biofilm evidence and lack of a single validated inorganic system combining both stealth and biofilm-engaging behavior &
        Retained: redox-sheddable stealth shell exposing a biofilm-binding or antimicrobial surface. \newline Blueprinted: carbohydrate-coated Au--Ag nanoparticle with an added antifouling overlayer &
        \texttt{future neighborhood}; nearest future match was an engineered mesoporous silica antimicrobial nanoparticle paper \\
        \midrule
        C &
        \texttt{cluster pair 0-13} \newline 64-paper pack &
        Therapeutic nanocarriers and delivery coatings versus quantum-dot imaging, passivation, and bioconjugation literatures &
        Strong delivery/imaging support (\(\approx 0.93\)), but the auditor still judged the biofilm-overcoming claim indirect and requested evidence about charge, hydrophobicity, enzyme responsiveness, and matrix disruption &
        Retained: stimulus-responsive polymer coating on a fluorescent inorganic nanoparticle. \newline Blueprinted: enzyme-cleavable coating that shrinks an inorganic particle cluster after trigger exposure &
        \texttt{future neighborhood}; nearest future match was a magneto-fluorescent mesoporous antimicrobial nanocarrier paper \\
        \bottomrule
    \end{tabular}
\end{table*}

\subsection{Example A: cluster-pair bridge between responsive delivery and multifunctional inorganic platforms}

\paragraph{Evidence pack.}
The \texttt{cluster\_pair\_0\_3} example is a good illustration of how \textit{pArticleMap} builds a deliberately mixed pack rather than a keyword-only bundle. The exported evidence pack contains 64 papers and combines cue-hit papers, cluster exemplars, and boundary papers. The top retrieved items included PNIPAM-coated nanostructures, a CD47-peptide anti-phagocytic drug nanocarrier, a review of reactive-oxygen-species-responsive drug delivery systems, an erythrocyte-camouflaged hollow copper-sulfide nanoplatform, and a pH-triggered radioluminescent nanocapsule study. In other words, the pack splices together stealth-coating logic, triggerable release logic, and multifunctional inorganic theranostics before any hypothesis is written.

\paragraph{Explanation.}
The explanation agent summarized cluster A as a literature centered on stimulus-responsive nanoparticle delivery, especially ROS/redox-triggered release, and cluster B as a literature centered on multifunctional tumor-therapeutic platforms emphasizing photothermal or imaging functions together with circulation-improving coatings. The bridge opportunity was therefore not framed as ``biofilm nanoparticles'' in a generic sense, but more specifically as the transfer of triggerable surface-control strategies from delivery systems into coated inorganic platforms.

\paragraph{Audit and patch retrieval.}
The auditor still refused to treat that bridge as fully sufficient evidence for antibiofilm design. Its stored support fraction was approximately \(0.96\), yet the same audit listed missing evidence on extracellular-polymeric-substance penetration, coating chemistries suitable for inorganic biofilm use, whether biofilms actually provide the relevant ROS or redox trigger, and basic safety constraints for such coatings. The corresponding patch queries explicitly asked for inorganic nanoparticle coatings for biofilm penetration, EPS-disruptive or cationic coatings, and biofilm microenvironment triggers. This example shows the central conservatism of the system: a coherent frontier explanation is not automatically accepted as adequate support for a cue-specific mechanistic claim.

\paragraph{Ideation, scoring, and blueprinting.}
After retrieval repair, the ideation agent produced five substantially different candidate bridges. The review packet retained the hypothesis \emph{``A charge-reversible, pH-sheddable coating on inorganic nanoparticles will balance transport and adhesion better than constitutively charged surfaces in biofilm models.''} The reviewer summary described it as the strongest hypothesis in that set on experimental clarity and actionability, but still dependent on an uncertain biofilm pH trigger. Internally, however, the blueprint agent did not expand that hypothesis. The highest-scored idea in the same trace was instead a polydopamine-coated photothermal inorganic nanoparticle with ROS/pH/NIR-triggered drug release. Its blueprint translated the idea into a bill of materials centered on a Prussian-blue or \(\mathrm{Fe_3O_4@Prussian\ blue}\) core, a polydopamine interfacial shell, an antimicrobial payload, NIR activation, and success criteria tied to loading stability, triggered release, photothermal heating, and biofilm biomass reduction. This divergence is useful: retrospective matching and internal experimental prioritization are related but not identical selection problems.

\paragraph{Retrospective interpretation.}
For the retrospective benchmark, this trace yielded a \texttt{future neighborhood} outcome rather than an exact gold recovery. The closest future neighbor was a pH-triggered antibacterial silver nanoparticle paper, which indicates that the generated idea reached the correct surface-engineering neighborhood without yet anchoring strongly enough to the exact later paper chosen as gold.

\subsection{Example B: broad gap-region bridge between stealth coatings and biofilm-engaging surfaces}

\paragraph{Evidence pack.}
The \texttt{gap\_0} example shows how \textit{pArticleMap} behaves in a broad sparse region rather than at a clean cluster interface. Its 69-paper evidence pack drew from cue-relevant papers, gap-boundary papers, and exemplars from several touched communities. The two sides of the gap were explained as one literature on anti-clearance or circulation-preserving coatings and another on targeted, adhesive, triggerable, or antimicrobial surface modifications. This is precisely the kind of fragmented design space that is difficult to reach with ordinary keyword searches, because the concepts are adjacent in function but dispersed across distinct literatures.

\paragraph{Audit and patch retrieval.}
This example received the harshest audit among the three worked cases, with a support fraction of approximately \(0.79\). The auditor explicitly flagged four gaps: no direct evidence that the cited coatings penetrate mature biofilms, no direct evidence that the coatings were optimized for biofilm environments rather than cancer or general delivery, no clear articulation of the stealth-versus-adhesion tradeoff required for biofilms, and no single validated inorganic system unifying both behaviors. The patch queries therefore targeted dual-function coating studies, cationic or matrix-binding coatings, and stimuli-responsive inorganic nanoparticle coatings for biofilm eradication.

\paragraph{Ideation, scoring, and blueprinting.}
The review packet retained the hypothesis \emph{``Redox-sheddable stealth coating on inorganic nanoparticles will improve serum persistence yet expose a biofilm-binding/antimicrobial surface at the infection site.''} The reviewer summary judged it to be a reasonable bridge with clear experiments, but not yet well grounded in biofilm-specific functional evidence. Within the same runtime trace, the blueprint agent instead selected a different candidate: a carbohydrate-coated bimetallic Au--Ag nanoparticle with an added antifouling overlayer. The stored blueprint operationalized that design into a concrete preclinical plan with serum stability and protein-adsorption assays, macrophage uptake measurements, planktonic and biofilm-killing readouts, and an infected-wound or local biofilm model as the translational endpoint. This is a useful example of \textit{pArticleMap} preferring a more operationally closed design even when the later review packet highlights a different hypothesis as the most relevant future-neighbor candidate.

\paragraph{Retrospective interpretation.}
The retained hypothesis again landed in \texttt{future neighborhood}. Its nearest future neighbor was an engineered mesoporous silica antimicrobial nanoparticle study, while the strongest historical confound involved stealth iron-oxide nanoparticles. In practice, this means \textit{pArticleMap} was able to connect stealth-surface logic with infection-oriented inorganic platforms, but it had not yet sufficiently isolated what is uniquely required for mature biofilm penetration and retention.

\subsection{Example C: under-supported bridge from imaging passivation to biofilm-active coatings}

\paragraph{Evidence pack.}
The \texttt{cluster\_pair\_0\_13} case is the clearest example of a scientifically plausible but under-supported bridge. The explanation agent contrasted a drug-delivery literature focused on loading, release, targeting, and circulation against a quantum-dot and inorganic imaging literature dominated by passivation, water solubilization, and bioconjugation. The pack therefore made it natural to ask whether an imaging-oriented passivation strategy could be converted into a biofilm-active coating strategy on fluorescent inorganic particles.

\paragraph{Audit and patch retrieval.}
The audit support fraction remained high at approximately \(0.93\), but the missing-facet diagnosis is more important than the headline number. The auditor wrote that the pack still lacked biofilm-specific coating requirements, lacked mechanistic attributes for penetrating or disrupting biofilms such as charge, hydrophobicity, enzyme responsiveness, or matrix-degrading functionality, and lacked direct evidence that the cited coatings help inorganic particles overcome biofilms rather than merely remain stable and imageable. This is exactly the sort of case in which a naive generator would likely overclaim novelty, whereas the audited framework continues to mark the bridge as conditional.

\paragraph{Ideation, scoring, and blueprinting.}
The review packet retained the hypothesis \emph{``Stimulus-responsive polymer coatings on fluorescent inorganic nanoparticles will outperform passive passivation in biofilm drug release while preserving optical tracking.''} The reviewer described it as actionable and testable but under-supported for its core biofilm-overcoming claim. The blueprint agent again selected a different candidate from the same trace: an enzyme-cleavable coating that holds inorganic particles in a larger imaging-capable cluster before trigger-dependent disassembly into smaller penetrant fragments. The stored blueprint demanded trigger-dependent size reduction, fluorescence recovery after cleavage, deeper biofilm penetration than non-cleavable controls, and improved antimicrobial effect after shrinkage. This is an informative behavioral pattern: once the audit establishes that simple passivation is too generic, the higher-ranked ideas tend to migrate toward multistage or transformable designs rather than static coatings.

\paragraph{Retrospective interpretation.}
This trace also ended in \texttt{future neighborhood}, with the closest future neighbor being a magneto-fluorescent mesoporous nanocarrier for dual antimicrobial delivery. The result is qualitatively sensible: \textit{pArticleMap} anticipated the convergence of imaging functionality and antimicrobial transport, but the available historical evidence was still too indirect to justify a stronger claim about biofilm-specific coating principles.

\subsection{Behavioral Takeaways}

Taken together, these examples illustrate five recurring properties of \textit{pArticleMap}.

\begin{itemize}
    \item \textbf{The evidence pack is a designed object, not a bag of hits.} Each pack mixes cue-responsive retrieval, cluster exemplars, and boundary papers so that the model sees both sides of a frontier and the plausible bridge region between them.
    \item \textbf{The auditor is the main conservatism mechanism.} Even when explanations are locally well supported, the system explicitly records missing cue-specific evidence such as mature-biofilm data, matrix-penetration mechanisms, or trigger realism.
    \item \textbf{Ideation is plural rather than singular.} A single trace typically yields multiple bridge families, including pH-sheddable coatings, biomimetic anti-clearance shells, polydopamine trigger layers, transformable clusters, and imaging-readable release coatings.
    \item \textbf{Blueprinting follows internal scoring, not retrospective hindsight.} The blueprint agent expands the highest-scored experimentally actionable idea from a trace, whereas the review packet later retains the hypothesis that best matches a held-out future-paper task. Those two choices often differ, and that difference is scientifically useful.
    \item \textbf{The antimicrobial run behaves consistently with the main text.} The examples are highly testable and often experimentally clear, but many remain only directionally grounded in biofilm mechanisms. This is the appendix-level qualitative explanation for why Section~\ref{results} reports relatively strong feasibility and evaluability but weaker novelty and plausibility in the first antimicrobial retrospective run.
\end{itemize}

\section{Consensus-Top Vaccine Example from Human Review}\label{appendix:worked-vaccine}

We also inspected the vaccine assessment bundle, focusing on a specific idea. Under the current human-analysis outputs, this is the highest-mean vaccine example. The originating cue was \emph{``What adjuvants can be included in mRNA LNP vaccines to improve their long term efficacy?''}, and the effective target was \texttt{cluster pair 0-11} sourced from \texttt{gap 4}.

\begin{rhoenv}[frametitle=Full Evidence-Pack Excerpt]
The published evidence pack contained 64 papers. The generated hypothesis directly cited the following support papers:
\begin{itemize}
    \item \texttt{id:27105589\_src18536}: \emph{Supramolecular Hydrogel from Nanoparticles and Cyclodextrins for Local and Sustained Nanoparticle Delivery} (2016; \texttt{cluster\_11\_boundary}).
    \item \texttt{id:31389136\_src31859}: \emph{Site-Specific Construction of Long-Term Drug Depot for Suppression of Tumor Recurrence} (2019; \texttt{discovery\_cue\_query}).
    \item \texttt{id:24070571\_src10970}: \emph{A magnetic chitosan hydrogel for sustained and prolonged delivery of Bacillus Calmette--Gu\'erin in the treatment of bladder cancer} (2013; \texttt{cluster\_0\_boundary}).
    \item \texttt{id:29246871\_src22820}: \emph{Injectable methylcellulose hydrogel containing calcium phosphate nanoparticles for bone regeneration} (2018; \texttt{cluster\_11\_exemplar}).
    \item \texttt{id:29957531\_src26244}: \emph{Incorporation of BMP-2 nanoparticles on the surface of a 3D-printed hydroxyapatite scaffold using an $\varepsilon$-polycaprolactone polymer emulsion coating method for bone tissue engineering} (2018; \texttt{cluster\_11\_exemplar}).
\end{itemize}
Taken together, these papers bridge sustained local depot formation, prolonged nanoparticle residence, and scaffold-associated release systems. The explain stage therefore surfaced a biomaterial-inspired mRNA-LNP depot bridge, but it still marked the pack as insufficient evidence for direct claims about vaccine adjuvant inclusion and long-term efficacy.
\end{rhoenv}

\begin{rhoenv}[frametitle=Full Generated Hypothesis]
\textbf{Title.} \emph{Injectable supramolecular hydrogel depots that locally retain mRNA-LNPs and a Th1-biasing immunostimulant will improve durability of vaccine responses versus bolus LNP injection}. \\
\textbf{Text.} Cluster A evidence shows that local depot formation and sustained release can extend residence and improve efficacy of payloads: a supramolecular hydrogel released discrete nanoparticles over $\sim$2 weeks after subcutaneous injection and was biocompatible (\texttt{id:27105589\_src18536}), while in situ fibrous depots enabled long-term local drug retention and sustained activity (\texttt{id:31389136\_src31859}). A magnetic chitosan hydrogel prolonged local residence of BCG and increased Th1 immune responses and antitumor efficacy relative to conventional delivery (\texttt{id:24070571\_src10970}). Bridging to Cluster B, injectable hydrogels and scaffold-associated nanoparticles support sustained local release of bioactives and improved tissue responses (\texttt{id:29246871\_src22820}; \texttt{id:29957531\_src26244}). A testable hypothesis is that embedding mRNA-LNPs plus an added immunostimulant in a shear-thinning hydrogel depot will prolong local antigen/adjuvant exposure, enhance lymphatic drainage over time, and produce stronger memory responses than free LNP. \\
\textbf{Support citations.} \texttt{id:27105589\_src18536}, \texttt{id:31389136\_src31859}, \texttt{id:24070571\_src10970}, \texttt{id:29246871\_src22820}, and \texttt{id:29957531\_src26244}.
\end{rhoenv}

\paragraph{Human review summary.}
Current human-analysis outputs treat this as the consensus-top reviewed vaccine example. Mirre submitted criterion scores of importance \(=3\), novelty \(=4\), plausibility \(=4\), feasibility \(=3\), evaluability \(=4\), and likely impact \(=3\), while Koen recorded a scored draft with \(4\) on all six criteria. The matched means used in the paired analysis were therefore importance \(=3.5\), novelty \(=4.0\), plausibility \(=4.0\), feasibility \(=3.5\), evaluability \(=4.0\), and likely impact \(=3.5\), giving an overall human mean of \(3.75/5\).

\paragraph{Audit note.}
The audit reported a supported-claim fraction of approximately \(0.92\), but still set \texttt{needs\_patch = True} and the explain stage recorded \texttt{insufficient\_evidence = True}. It explicitly flagged missing direct evidence for mRNA-LNP vaccine adjuvants, long-term efficacy gains, suitable adjuvant classes, safety/reactogenicity trade-offs, and the practical inclusion route. Its bridge seed was \emph{``A biomaterial-inspired mRNA-LNP vaccine depot that combines sustained local release with immune-modulating components,''} accompanied by the warning that no evidence in the pack specifically addressed mRNA-LNP vaccines or long-term vaccine efficacy.

\section{Text Encoder Evaluation}
\label{appendix:text-encoder-eval}

\begin{figure}[t]
    \centering
    \includegraphics[width=0.8\linewidth]{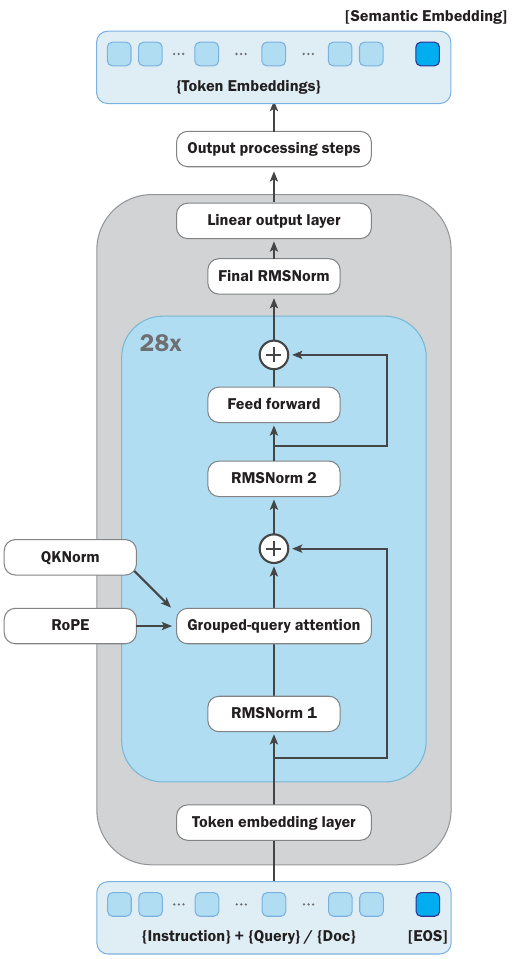}
    \caption{Qwen3 Embedding model architecture}
    \label{fig:qwen3}
\end{figure}

Selecting an appropriate document encoder is important for the downstream retrieval, clustering, and novelty-discovery stages of the pipeline. We therefore compared two candidate embedding models, \texttt{Qwen3-Embedding-0.6B} (Figure~\ref{fig:qwen3}) and \texttt{BioClinical ModernBERT-large}, using a shared evaluation protocol on the same cleaned corpus.
The input file contained 67,043 records. After removing documents with empty or placeholder MeSH annotations, 57,606 documents remained. Embeddings and metadata were kept strictly row-aligned, all vectors were finite and non-zero, and both embedding matrices had dimensionality 1,024. MeSH labels were lower-cased, de-duplicated per document, and filtered to retain only labels occurring at least five times. This yielded 5,605 corpus-level MeSH classes for retrieval, with a mean of 13.06 retained MeSH labels per document.

\subsection{Evaluation protocol}

We used two complementary evaluations to compare the embedding spaces.

\paragraph{Linear probing.}
We first measured how linearly accessible the MeSH label structure was from each embedding space. For each model, we created a single 80/20 train/test split, yielding 46,084 training and 11,522 test documents. The probe label space was defined only from the training split using the same minimum-frequency threshold, resulting in 5,058 train-supported classes. We trained a single linear multilabel classifier with binary cross-entropy on CUDA for 20 epochs using AdamW, batch size 512, learning rate $10^{-3}$, weight decay $10^{-4}$, and positive-class weights clipped at 100. Threshold tuning was disabled, and predictions were obtained with a fixed threshold of 0.5.

\paragraph{Nearest-neighbour retrieval.}
We also evaluated each encoder using unsupervised cosine-similarity retrieval over normalized embeddings. A query--document pair was treated as binary-relevant when the two documents shared at least one filtered MeSH label, while graded relevance was defined as the number of shared labels. Retrieval metrics were averaged over 5,000 sampled queries. For reference, we also included a lexical TF--IDF baseline built from title and abstract text.

\begin{table*}[t]
\centering
\small
\caption{Embedding model comparison on MeSH linear probing and keyword-overlap retrieval. Bold values indicate the better of the two embedding models. LRAP denotes label-ranking average precision. Higher values are better for Micro-F1, Macro-F1, LRAP, MRR, P@10, MAP@10, nDCG@10, and Shared@10; lower values are better for ranking loss and Hamming loss.}
\label{tab:embedding_model_selection}
\begin{tabular}{lcccccccccc}
\toprule
& \multicolumn{5}{c}{Linear probe} & \multicolumn{5}{c}{Retrieval} \\
\cmidrule(lr){2-6} \cmidrule(lr){7-11}
Model / baseline
& F1$_\mu\uparrow$ & F1$_M\uparrow$ & LRAP$\uparrow$ & RL$\downarrow$ & Ham.$\downarrow$
& MRR$\uparrow$ & P@10$\uparrow$ & MAP@10$\uparrow$ & nDCG@10$\uparrow$ & Shared@10$\uparrow$ \\
\midrule
Label-prior baseline
& 0.304 & 0.001 & 0.296 & 0.0756 & 0.0036
& -- & -- & -- & -- & -- \\
TF--IDF baseline
& -- & -- & -- & -- & --
& 0.994 & 0.982 & 0.973 & 0.243 & 4.306 \\
BioClinical ModernBERT-large
& 0.176 & 0.116 & 0.383 & 0.0453 & 0.0186
& 0.979 & 0.953 & 0.929 & 0.126 & 3.438 \\
Qwen3-Embedding-0.6B
& \textbf{0.212} & \textbf{0.182} & \textbf{0.466} & \textbf{0.0396} & \textbf{0.0154}
& \textbf{0.996} & \textbf{0.988} & \textbf{0.981} & \textbf{0.267} & \textbf{4.564} \\
\bottomrule
\end{tabular}
\end{table*}

\subsection{Results and model choice}

\texttt{Qwen3-Embedding-0.6B} outperformed \texttt{BioClinical ModernBERT-large} across all embedding-based metrics. In the linear-probe setting, Qwen improved micro-F1 from 0.176 to 0.212, macro-F1 from 0.116 to 0.182, and LRAP from 0.383 to 0.466, while also reducing ranking loss and Hamming loss. The gain in macro-F1 is particularly important because the MeSH prediction problem is highly imbalanced, indicating better recovery of less frequent labels rather than only improved prediction of common terms.

The label-prior baseline achieved a higher micro-F1 than either embedding probe, but this mainly reflects the extreme frequency skew of MeSH annotations: it predicts the same most frequent labels for every document. Its macro-F1 was only 0.001, and its LRAP was substantially below both embedding-based probes. For model selection, we therefore regarded macro-F1, LRAP, and retrieval quality as more informative than the thresholded micro-F1 of the prior baseline.

The retrieval results showed the same pattern. Qwen achieved higher MRR, Precision@10, MAP@10, nDCG@10, and mean shared MeSH labels among the top retrieved documents. At $k=10$, it increased MAP from 0.929 to 0.981 and improved nDCG@10 from 0.126 to 0.267. The mean number of shared MeSH labels among the top 10 neighbours increased from 3.438 to 4.564, indicating that Qwen retrieved not only documents with at least one overlapping MeSH term, but documents with stronger graded topical overlap.

Binary retrieval metrics should be interpreted with caution because the relevance definition is broad: on average, each sampled query had more than 50,000 documents sharing at least one filtered MeSH label. As a result, precision- and hit-based metrics are close to saturation. The graded retrieval measures, especially nDCG@10 and mean shared labels, are more discriminative in this setting. On these measures, Qwen was consistently stronger than both ModernBERT and the TF--IDF baseline.

Based on these results, we selected \texttt{Qwen3-Embedding-0.6B} as the document encoder for the downstream analyses. Its advantage was consistent across both supervised probing and unsupervised retrieval, making it the most suitable embedding model for the final pipeline.

\section{Implementation and Reproducibility Details}
\label{appendix:technical-details}

This appendix records the concrete implementation details available from the repository manifests, run packets, and local runtime configuration used for the reported \textit{pArticleMap} analyses. It is intended as a reproducibility supplement to the Methods and Experimental Setup sections rather than a replacement for the released code.

\subsection{Corpus, embeddings, and retrieval services}

The cleaned corpus used by the analysis pipeline contains 67,043 PubMed-derived records. Each model input was the concatenation of paper title and abstract, with bibliographic metadata, DOI when available, author keywords, MeSH terms, language fields, and publication-date components retained separately. The selected downstream document encoder was \texttt{Qwen/Qwen3-Embedding-0.6B}. The stored Qwen embedding manifest records 1,024-dimensional L2-normalized vectors, \texttt{float32} storage with no corpus-level instruction string. The comparison encoder used for model selection was \texttt{thomas-sounack/BioClinical-ModernBERT-large}, also stored as 1,024-dimensional normalized vectors.

At evaluation time, retrieval used a local FastAPI service exposing \texttt{/embed} and \texttt{/rank}. The service wrapped \texttt{Qwen/Qwen3-Embedding-0.6B} for embedding and \texttt{Qwen/Qwen3-Reranker-0.6B} for reranking. Both Qwen endpoints used a maximum sequence length of 8,192 tokens. On CUDA, the Qwen service used \texttt{float32} model weights unless overridden by environment variables. If reranking failed, the \texttt{/rank} endpoint fell back to embedding-similarity scores so that candidate ranking could still complete.

\begin{table*}[t]
\centering
\small
\caption{Model and runtime disclosure for the reported \textit{pArticleMap} pipeline.}
\label{tab:implementation_runtime_disclosure}
\begin{tabular}{p{0.20\textwidth}p{0.28\textwidth}p{0.42\textwidth}}
\toprule
\textbf{Component} & \textbf{Model or service} & \textbf{Configuration and role} \\
\midrule
Corpus encoder & \texttt{Qwen3-Embedding-0.6B} & Selected document encoder; 1,024-dimensional L2-normalized vectors; \texttt{float32} stored embeddings; 67,043 papers; no corpus-embedding instruction;  \\
\midrule
Comparison encoder & \texttt{BioClinical-ModernBERT-large} & Alternative encoder evaluated in Appendix~\ref{appendix:text-encoder-eval}; 1,024-dimensional normalized vectors; used for model selection rather than final downstream analysis. \\
\midrule
Retriever and reranker & Local FastAPI Qwen service with \texttt{/embed} and \texttt{/rank} & Embedding endpoint used Qwen3 embeddings; rank endpoint used \texttt{Qwen/Qwen3-Reranker-0.6B} plus optional embedding-similarity scores; maximum sequence length 8,192. \\
\midrule
Agent LLMs & OpenAI chat models & Reported retrospective runs passed \texttt{GPT 5.4-mini 2026-03-17} through \texttt{ OpenAI} for explanation, audit, blueprinting, and LLM judging. Ideation used the orchestrator-pinned \texttt{GPT 5.4 2026-03-05} with medium reasoning effort. \\
\midrule
Fallbacks & Deterministic baselines and heuristic scorer & The evaluation stack includes \texttt{heuristic bridge}, \texttt{pack query baseline}, and a heuristic idea scorer for runs without an available LLM judge. \\
\bottomrule
\end{tabular}
\end{table*}

\subsection{Snapshot analysis configuration}

The cue-conditioned retrospective bundles used Qwen embeddings as the analysis representation. PCA was enabled with 102 components before graph construction and clustering. The stored run manifests specify Leiden clustering with community resolution 1.0, community graph \(k=21\), cosine graph metric, analysis graph \(k=21\), cosine density metric, density scales \(\{10,20,30,40,50\}\), gap quantile 0.95, minimum gap-region size 3, UMAP neighbors 50, UMAP minimum distance 0.1, and random seed 42. UMAP coordinates were stored for visualization only and were not used for gap scoring, target construction, or recovery matching. The four reported cue-conditioned review packets share the analysis-configuration.

The published snapshot is the reproducibility boundary for downstream generation. Once a historical snapshot is stored, evidence-pack construction, target selection, generation, scoring, blueprinting, and retrospective matching operate against that snapshot identifier rather than mutable in-memory analysis state. The backend stores papers, embeddings, clusters, gap memberships, generated research briefs, evaluation runs, and match records in a SQLite knowledge store served through FastAPI.

\subsection{LLM prompting and evaluation settings}

The LangGraph orchestrator used structured state transitions for \texttt{build pack}, \texttt{explain}, \texttt{audit}, optional \texttt{patch\_retrieve}, \texttt{ideate}, \texttt{score}, \texttt{blueprint}, and \texttt{publish}. Structured outputs were enforced with Pydantic schemas through LangChain function-calling interfaces. The system prompts instructed the models to use only the provided evidence pack, cite by \texttt{paper id}, treat discovery cues as steering context rather than evidence, and mark unsupported details as unknown or assumptions. Non-judge generation used temperature 0.2 in the runtime configuration, while LLM judging used temperature 0.0.

The four cue-conditioned retrospective bundles used cutoff date \texttt{2019-12-31} and a future evaluation window from \texttt{2020-01-01} to \texttt{2026-01-01}. Each bundle ran the \texttt{orchestrator} method only, with 20 gap targets, 10 cluster-pair targets, 50 gold future papers, 1 seed, and 3 hypotheses per target. The future-paper pool was semantically prefiltered with threshold 0.45 for each domain cue. The stored manifests specify historical-confound retrieval and scoring for generated hypotheses during evaluation.

\subsection{Software environment and availability caveat}

The inspected runtime used Python 3.10.11. Dependencies are recorded as lower-bound requirement files rather than as an exact package lockfile. The main application stack used Streamlit for the analysis and assessment interfaces, FastAPI with Uvicorn for the backend and embedding services, SQLite for the agent knowledge store, PyTorch and Transformers for local embedding and reranking models, scikit-learn and NetworkX for PCA, neighbor graphs, and related analysis, and LangGraph/LangChain for agent orchestration. Optional tracing used Langfuse services configured through Docker Compose.


  \end{document}